\documentclass[3p,sort&compress]{elsarticle}
\usepackage{lmodern} 
\usepackage{microtype}

\usepackage[lined,ruled,linesnumbered,vlined]{algorithm2e}
\usepackage{amssymb}
\usepackage{amsmath}
\usepackage{graphicx}
\usepackage{booktabs}
\usepackage{multirow}
\usepackage{subcaption}
\usepackage{adjustbox}
\usepackage{enumitem}
\usepackage{mathrsfs}
\usepackage[colorlinks=true]{hyperref}

\begin{document}

\begin{frontmatter}
\title{Adaptive regularization parameter selection for high-dimensional inverse problems: A Bayesian approach with Tucker low-rank constraints}

\author{Qing-Mei Yang}
\author{Da-Qing Zhang\corref{cor1}}
\ead{d.q.zhang@ustl.edu.cn}
\cortext[cor1]{Corresponding author}	
\address{School of Science, University of Science and Technology Liaoning,
	Anshan, Liaoning Province, 114051, PR China}

\begin{abstract}
	This paper introduces a novel variational Bayesian method that integrates Tucker decomposition for efficient high-dimensional inverse problem solving. The method reduces computational complexity by transforming variational inference from a high-dimensional space to a lower-dimensional core tensor space via Tucker decomposition. A key innovation is the introduction of per-mode precision parameters, enabling adaptive regularization for anisotropic structures. For instance, in directional image deblurring, learned parameters align with physical anisotropy, applying stronger regularization to critical directions (e.g., row vs. column axes). The method further estimates noise levels from data, eliminating reliance on prior knowledge of noise parameters (unlike conventional benchmarks such as the discrepancy principle (DP)). Experimental evaluations across 2D deblurring, 3D heat conduction, and Fredholm integral equations demonstrate consistent improvements in quantitative metrics (PSNR, SSIM) and qualitative visualizations (error maps, precision parameter trends) compared to L-curve criterion, generalized cross-validation (GCV), unbiased predictive risk estimator (UPRE), and DP. The approach scales to problems with 110,000 variables and outperforms existing methods by 0.73–2.09 dB in deblurring tasks and 6.75 dB in 3D heat conduction. Limitations include sensitivity to rank selection in Tucker decomposition and the need for theoretical analysis. Future work will explore automated rank selection and theoretical guarantees. This method bridges Bayesian theory and scalable computation, offering practical solutions for large-scale inverse problems in imaging, remote sensing, and scientific computing.

\end{abstract}


\end{frontmatter}

\section{Introduction} 

High-dimensional inverse problems arise ubiquitously in medical imaging \cite{XJAZEB4BBB18022424582B54C1685B37205B,SJES04508D2FADF8516E6C39E9AD49261354}, signal processing \cite{Knobles2018SignalAI,SIJD15112600080412}, and remote sensing \cite{SJWD25A2C8217DDB02D0CCE32FF15771B118}. These problems can be formulated as solving a linear system $y = Ax + \epsilon$, where $A \in \mathbb{R}^{m \times n}$ denotes the forward operator, $y \in \mathbb{R}^{m}$ the observation vector, $\epsilon \sim \mathcal{N}(0, \sigma^2 I)$ Gaussian noise, and $x \in \mathbb{R}^{n}$ the unknown to be estimated. Ill-posedness or under-determinacy fundamentally challenges these inverse problems by jeopardizing the existence, uniqueness, and stability of solutions \cite{NSTLCF7D816C3BBA34AF2DD788A69F5654BC}. Regularization methods address this difficulty by incorporating prior knowledge to constrain the admissible solutions. Tikhonov regularization \cite{SBIPDD88373FA38A12874838199DC76886A0,SBXQ7A157F666F171CF2965208029064D659} exemplifies this approach by appending a norm penalty to the least-squares cost function to enforce a compromise between data fidelity and solution smoothness. A key consideration in any regularization scheme is how to choose the parameter $\lambda$ that weights the regularization term against data fidelity.
Several methods exist for selecting regularization parameters. The L-curve method \cite{hansen1992lcurve} identifies the parameter at the point of maximum curvature in the log-residual versus log-solution-norm plot. This method is intuitive and noise-level independent but suffers from non-smooth curves \cite{SJES96DB7985172720646EFC2744A3C7E73C} and ambiguous corner localization \cite{SSJD00003358733} in high dimensions with sensitivity to problem conditioning. Generalized cross-validation (GCV) \cite{STJDF74788E9975C8E9A2B24D6E91637CC9EC} approximates leave-one-out cross-validation efficiently. Unbiased predictive risk estimation (UPRE) \cite{SJES8BD5EBB33395A742B4F6A6F5A2E03EF1} requires known noise variance and thus has limited practical applicability. The Morozov discrepancy principle \cite{STJDB2C9B99C810BB4D1D97ACC55A3E13E55} matches residuals to noise levels with theoretical guarantees but also requires accurate prior knowledge of $\sigma$.

Many practical problems involve multi-dimensional data that naturally take tensor form
\cite{SQGF1122B6B516B2BA5EEB23AD3650BFF282}. Dynamic medical images \cite{9632459} constitute 
three- or four-dimensional tensors indexed by time and spatial coordinates. Hyperspectral remote sensing 
data \cite{SJMD91C257D431A72A6FE3787A369D52CD3C} are organized as three-dimensional tensors 
over spatial location and spectral bands. Video sequences 
\cite{SJES4A1F35B147CB056E501001F91C8FB887} form third-order tensors spanning temporal and 
spatial dimensions. Such data commonly exhibit intrinsic low-rank structure that permits accurate 
approximation using relatively few parameters.
This low-rank property suggests a natural strategy for overcoming the dimensionality barrier in variational Bayesian inference. Constraining the unknowns to a low-dimensional subspace defined through Tucker structure \cite{SJESC32CAFE384C17D1EAB81FA04840B353B,SJES038207BA59A22B5EE728CC118698604E} reformulates the variational inference problem from the original $n$-dimensional parameter space to an $n_{\mathcal{G}}$-dimensional core tensor space. The posterior covariance matrix consequently reduces from $n \times n$ to $n_{\mathcal{G}} \times n_{\mathcal{G}}$, and computational complexity drops from $\mathcal{O}(n^3)$ to $\mathcal{O}(n_{\mathcal{G}}^3)$.
The Tucker structure here functions as a subspace constraint with factor matrices predetermined by the mathematical structure of the problem rather than learned from data. This subspace constraint is analogous to sparsity assumptions in compressed sensing; both serve as structured priors that regularize ill-posed inverse problems. Unlike diagonal approximations, the core tensor in Tucker decomposition remains dense and allows the rank along each mode to be specified independently. This mode-wise flexibility accommodates direction-dependent characteristics inherent in the data and provides the basis for differentiated regularization across tensor modes.

Regularization has a well-known probabilistic counterpart \cite{NSTLBB44C87B6FC91B5334E0EB48F33B1BF9}. The Bayesian framework models the unknown vector $x$ as a random variable and incorporates prior knowledge via specified prior distributions. When both the prior and the noise are assumed Gaussian, the posterior mean coincides with the Tikhonov solution and the regularization parameter $\lambda$ can be interpreted as the ratio of noise variance to signal variance .
Bayesian hierarchical models offer a principled route to automatic selection of regularization parameters. These models treat hyperparameters of the prior as random variables endowed with their own hyperpriors so that unknowns and hyperparameters are estimated jointly through posterior optimization. Empirical Bayes, a popular variant, determines hyperparameters by maximizing the marginal likelihood; this approach has proven effective in source localization \cite{Antoni2012,Pereira2015} and vibration load identification \cite{Coletti2025}. Bayesian optimization has likewise gained traction for hyperparameter tuning in machine learning \cite{SJPD20BFEC0F876EF9736AFDC4C0E58D585D}. A known drawback of empirical Bayes is that it may return zero as the optimal regularization parameter in underdetermined or square systems and lead to unstable solutions. Coletti et al. \cite{Coletti2025} introduced auxiliary noise-only measurements to avoid this pathology and ensure the estimated regularization parameter remains bounded away from zero.
Classical Tikhonov regularization uses a single scalar parameter. This assumption becomes restrictive when the data show direction-dependent sensitivities. Distributed Tikhonov regularization \cite{Calvetti2025} relaxes this restriction by replacing the scalar $\lambda$ with a vector $\boldsymbol{\theta} = (\theta_1, \ldots, \theta_p)^{\mathrm{T}}$ so that each component or group of components receives its own penalty weight. Such a formulation naturally accommodates anisotropic inverse problems.

Variational Bayesian inference \cite{XJAZ403838950E5E65397266212F045F64F5} offers a practical way to approximate complex Bayesian models. In contrast to empirical Bayes, which returns point estimates of hyperparameters, variational approaches construct a surrogate distribution $q(x,\lambda)$ to approximate the true posterior $p(x, \lambda | y)$; the approximation is refined by minimizing the Kullback--Leibler divergence \cite{XJAZ6AE7CF74C46B03D5DEBA10A4C37CCB90}. The mean-field assumption provides a common simplification under which the variational distribution factorizes into marginals and thereby converts high-dimensional integration into a sequence of tractable low-dimensional updates.
Applications of variational Bayesian methods to automatic regularization parameter selection span image restoration \cite{SJPDFCDA55B3E7C51996A19E1C38705C238A}, image segmentation \cite{SQGFBBD8CBC3A725B52C400E903D283EA175}, and other inverse problems. Variational updates have also been combined with evolutionary optimization for parameter inference in ordinary differential equation models arising in ecological networks \cite{SJWDEA7AD4A3AF69DFB707E1A3B3861299D0}.
Despite much progress on automatic regularization, nearly all existing methods operate in the original parameter space. This direct approach becomes computationally intractable in high-dimensional settings and precludes application to large-scale problems. The distributed regularization framework of Calvetti and Somersalo \cite{Calvetti2025} does incorporate hierarchical parameters, but iterating in the native space remains expensive when the unknown possesses tensor structure.

We develop an anisotropic Bayesian method for adaptive regularization parameter selection that exploits low-rank subspace constraints to address the scalability difficulties faced by variational Bayesian approaches in high-dimensional inverse problems. The underlying premise is that the unknown $x \in R^n$ admits accurate approximation within a predetermined low-dimensional Tucker subspace; variational inference in this reduced space circumvents the dimensional barrier. Projecting the inference onto the Tucker subspace yields exponential savings in computation while retaining the ability to estimate hyperparameters jointly. We further introduce a multi-hyperparameter prior that assigns an independent precision parameter $\{\alpha_k\}_{k=1}^d$ to each tensor mode and allows the regularization to adapt automatically to anisotropic problem structure.

The main contributions are summarized below.

\textbf{(1) Scalability.} Low-rank subspace constraints reduce the computational complexity of variational Bayesian inference from $\mathcal{O}(n^3)$ to $\mathcal{O}(n_{\mathcal{G}}^3)$ and memory requirements from $\mathcal{O}(n^2)$ to $\mathcal{O}(n_{\mathcal{G}}^2)$. This reduction makes variational Bayesian inference feasible for large-scale problems.

\textbf{(2) Anisotropic regularization.} We derive closed-form variational updates in the Tucker subspace rigorously. The multi-hyperparameter prior learns a separate regularization strength for each mode directly from the data and captures direction-dependent behavior without manual tuning.

\textbf{(3) Experimental evidence.} Four groups of experiments assess computational efficiency, reconstruction accuracy, noise estimation, and adaptability across problem types: algorithm timing tests, two-dimensional Fredholm integral equations, anisotropic image deblurring, and three-dimensional heat conduction inversion.

The paper proceeds as follows. Section 2 collects background material on high-dimensional inverse problems, Tikhonov regularization, and low-rank tensor representations. This section explains how Tucker structure can reformulate a high-dimensional problem in a much smaller space.
Section 3 builds the Bayesian model under subspace constraints. We begin with the likelihood and prior specification and then derive variational updates in the reduced Tucker space to obtain closed-form expressions for the posterior of the core tensor and the precision hyperparameters. The computational savings are quantified and we describe an anisotropic prior that assigns a separate hyperparameter to each tensor mode. Section 4 reports numerical experiments organized into four parts: (i) timing and scalability comparisons between Tucker-VB and direct VB; (ii) two-dimensional Fredholm integral equations of the first kind under varying noise levels benchmarked against five alternative parameter selection methods; (iii) anisotropic image deblurring with spatially varying kernels; and (iv) three-dimensional heat conduction inversion illustrating the method's capacity to recover direction-dependent structure automatically. Concluding remarks together with a discussion of limitations and possible extensions appear in Section 5.

\section{Preliminaries}
\subsection{High-Dimensional Inverse Problems and Regularization} 
Consider the discretized linear inverse problem $y = Ax + \epsilon$. Two issues frequently complicate the solution in practice. The system is underdetermined if the number of observations $m$ falls well below the unknown dimension $n$ ($m \ll n$). Even when $m \geq n$, a large condition number $\kappa(A) = \|A\| \cdot \|A^{-1}\|$ may still render the solution sensitive to noise.

Tikhonov regularization handles these difficulties by appending a penalty to the least-squares objective:
\begin{equation*}
	\hat{x} = \arg\min_{x} \left\{ \|y - Ax\|_2^2 + \lambda \|x\|_2^2 \right\}
\end{equation*}
whose minimizer is
\begin{equation*}
	\hat{x} = (A^T A + \lambda I)^{-1} A^T y
\end{equation*}
Here $\lambda > 0$ trades data fidelity against solution smoothness. Setting $\lambda$ too small yields a noise-sensitive estimate close to ordinary least squares. Setting it too large over-shrinks the solution and discards signal. Choosing $\lambda$ well is therefore central to obtaining useful reconstructions. The theory and algorithms for regularization are well developed when $n$ is moderate. As $n$ climbs into the $10^4$--$10^6$ range, variational Bayesian inference must form and store a posterior covariance $\Sigma_x \in \mathbb{R}^{n \times n}$. This demands $O(n^2)$ memory and $O(n^3)$ operations. The practical remedy is to exploit structure present in high-dimensional data.

\subsection{Low-Rank Tucker Prior}

The unknown vector $x \in R^{n}$ can often be reshaped into a $d$th-order tensor $\mathcal{X} \in R^{I_{1}\times I_{2}\times \cdots \times I_{d}}$, where $I_{k}$ is the size of the $k$th mode and $n = \prod_{k=1}^{d} I_k$. Under a low-rank Tucker prior, $\mathcal{X}$ is assumed to admit the multilinear decomposition
\begin{equation*}
	\mathcal{X} \approx \mathcal{G} \times_{1} U_{1} \times_{2} \cdots \times_{d} U_{d}
\end{equation*}
or in element-wise form
\begin{equation*}
	\mathcal{X}_{i_{1}, i_{2}, \ldots, i_{d}} \approx 
	\sum_{r_{1}=1}^{R_{1}} \sum_{r_{2}=1}^{R_{2}} \cdots \sum_{r_{d}=1}^{R_{d}} 
	\mathcal{G}_{r_{1}, r_{2}, \ldots, r_{d}} \prod_{k=1}^{d} (U_{k})_{i_{k}, r_{k}}
\end{equation*}
Here $\mathcal{G} \in R^{R_1 \times R_2 \times \cdots \times R_d}$ is the core tensor, $U_k \in R^{I_k \times R_k}$ $(k = 1, \ldots, d)$ are factor matrices, and $(R_1, R_2, \ldots, R_d)$ are multilinear ranks with $R_k \ll I_k$. The mode-$k$ product $\times_{k}$ is defined by
\begin{equation*}
	(\mathcal{Y} \times_{k} M)_{i_{1} \cdots i_{k-1}, j, i_{k+1} \cdots i_{d}} 
	= \sum_{i_{k}=1}^{I_{k}} \mathcal{Y}_{i_{1} \cdots i_{k} \cdots i_{d}} M_{j, i_{k}}
\end{equation*}
Geometrically, the columns of each $U_k$ form an orthonormal basis for mode $k$ and the core tensor $\mathcal{G}$ stores the coordinates of the solution in these bases. The parameterization reduces the degrees of freedom from $\prod_{k=1}^{d} I_{k}$ to $\prod_{k=1}^{d} R_{k} + \sum_{k=1}^{d} I_{k} R_{k}$. This represents an exponential saving when every $R_k$ is small.

For uniqueness and numerical stability, the factor matrices are taken to satisfy
\begin{equation*}
	U_k^\top U_k = I_{R_k}, \quad k = 1, \ldots, d
\end{equation*}
where $I_{R_k}$ is the $R_k \times R_k$ identity. The low-rank Tucker prior posits that the information content of $\mathcal{X}$ is concentrated in the small core tensor $\mathcal{G}$. Because $x$ is unknown and cannot be decomposed directly, the factor matrices must be fixed from the problem structure rather than learned from data. Only $\mathcal{G}$ remains as the Bayesian unknown. An attractive feature of this multilinear representation is that the $d$ modes of the core tensor remain separable with each carrying its own physical interpretation. This separation opens the door to anisotropic regularization: one can assign independent precision hyperparameters $\{\alpha_k\}_{k=1}^{d}$ to different modes.

To write the Tucker prior in vector notation, let $\boldsymbol{g} = \mathrm{vec}(\mathcal{G}) \in \mathbb{R}^{n_{\mathcal{G}}}$ with $n_{\mathcal{G}} = \prod_{k=1}^{d} R_k$. The subspace constraint then takes the matrix-vector form~\cite{SJES979C35ACACA8723F5CDB6270E9D6E5E2}
\begin{equation*}
	x = \tilde{U} \boldsymbol{g}
\end{equation*}
where $\tilde{U} = U_{d} \otimes U_{d-1} \otimes \cdots \otimes U_{1} \in R^{n \times n_{\mathcal{G}}}$ is the Kronecker product of all factor matrices. Substituting into the observation equation gives the reduced system
\begin{equation*}
	y = Ax + \epsilon = A\tilde{U} \boldsymbol{g} + \epsilon = \tilde{A} \boldsymbol{g} + \epsilon
\end{equation*}
with effective coefficient matrix $\tilde{A} = A\tilde{U} \in R^{m \times n_{\mathcal{G}}}$. Both computation and storage drop substantially because $n_{\mathcal{G}} \ll n$. Table~\ref{tab:complexity} summarizes the complexity comparison between the direct variational Bayes method and the proposed Tucker-VB method.

\begin{table}[!ht]
	\centering
	\caption{Complexity comparison}
	\label{tab:complexity}
	\begin{tabular}{l c c}
		\toprule
		& Direct VB & Tucker-VB \\ \midrule
		Posterior covariance size & $n \times n$ & $n_{\mathcal{G}} \times n_{\mathcal{G}}$ \\
		Matrix inversion          & $O(n^3)$     & $O(n_{\mathcal{G}}^3)$ \\
		Storage                   & $O(n^2)$     & $O(n_{\mathcal{G}}^2)$ \\
		Per-iteration cost        & $O(n^3 + mn^2)$ & $O(n_{\mathcal{G}}^3 + mn_{\mathcal{G}}^2)$ \\
		\bottomrule
	\end{tabular}
\end{table}

\section{Bayesian Model with Low-Rank Tucker Prior}
\subsection{Bayesian Model Formulation}
The Bayesian framework provides a probabilistic foundation for parameter estimation in inverse problems. Within this framework, all unknown quantities are treated as random variables. The prior distribution encodes existing knowledge, while the posterior distribution is derived from observed data through Bayes's theorem. Within this framework, the regularization term naturally corresponds to the introduction of a prior distribution. Based on the low-rank Tucker prior established in Section 2, we constructed a Bayesian model. Let $\beta$ denote the noise precision and assume Gaussian observation noise. The conditional probability density is:
\begin{equation*}
	p(y|\boldsymbol{g}, \beta) 
	= \mathcal{N}(y|\tilde{A}\boldsymbol{g}, \beta^{-1}I) 
	= \left(\frac{\beta}{2\pi}\right)^{m/2} 
	\exp\left(-\frac{\beta}{2} \|y - \tilde{A}\boldsymbol{g}\|_{2}^{2}\right)
\end{equation*}
where $\boldsymbol{g}$ collects the core tensor coefficients. We place a zero-mean Gaussian prior on $\boldsymbol{g}$, controlled by precision $\alpha$:
\begin{equation*}
	p(\boldsymbol{g}|\alpha) 
	= \mathcal{N}(\boldsymbol{g}|0, \alpha^{-1}I) 
	= \left(\frac{\alpha}{2\pi}\right)^{n_{\mathcal{G}}/2} 
	\exp\left(-\frac{\alpha}{2} \|\boldsymbol{g}\|_{2}^{2}\right)
\end{equation*}
The posterior then becomes
\begin{equation*}
	p(\boldsymbol{g}|y, \alpha, \beta) \propto 
	\exp\left\{-\frac{\beta}{2}\|y - \tilde{A}\boldsymbol{g}\|_{2}^{2} 
	- \frac{\alpha}{2}\|\boldsymbol{g}\|_{2}^{2}\right\}
\end{equation*}
and the MAP estimate solves
\begin{equation*}
	\hat{\boldsymbol{g}} = \arg\min_{\boldsymbol{g}} 
	\left\{\frac{\beta}{2}\|y - \tilde{A}\boldsymbol{g}\|_{2}^{2} 
	+ \frac{\alpha}{2}\|\boldsymbol{g}\|_{2}^{2}\right\}
\end{equation*}
Setting $\lambda = \alpha/\beta$ recovers the usual regularization parameter. The interpretation is straightforward: when measurements are noisy, we need more regularization. The reconstructed signal is $\hat{x} = \tilde{U}\hat{\boldsymbol{g}}$. In practice, $\alpha$ is often done by cross-validation or manual tuning, neither of which is entirely satisfactory since the optimal value depends on the unknown signal. A cleaner solution is to treat $\alpha$ and $\beta$ themselves as random variables and let the data determine appropriate values. Gamma priors are a natural choice here because of conjugacy:
\begin{equation*}
	p(\alpha) = \mathrm{Gamma}(\alpha|a_{\alpha}, b_{\alpha}) 
	= \frac{b_{\alpha}^{a_{\alpha}}}{\Gamma(a_{\alpha})} 
	\alpha^{a_{\alpha}-1} e^{-b_{\alpha}\alpha}
\end{equation*}
\begin{equation*}
	p(\beta) = \mathrm{Gamma}(\beta|a_{\beta}, b_{\beta}) 
	= \frac{b_{\beta}^{a_{\beta}}}{\Gamma(a_{\beta})} 
	\beta^{a_{\beta}-1} e^{-b_{\beta}\beta}
\end{equation*}
with shape parameters $a_{\alpha}, a_{\beta} > 0$ and rate parameters $b_{\alpha}, b_{\beta} > 0$. The joint posterior over all unknowns is
\begin{equation*}
	p(\boldsymbol{g}, \alpha, \beta | y) \propto 
	p(y|\boldsymbol{g}, \beta) \, p(\boldsymbol{g}|\alpha) \, p(\alpha) \, p(\beta)
\end{equation*}
Computing this posterior exactly is not feasible when dimensionality is high. We therefore turn to variational methods in the next subsection.

\subsection{Variational Bayesian Inference}

Variational Bayesian inference approximates the intractable true posterior $p(\mathcal{G}, \alpha, \beta | y)$ by a tractable distribution $q(\mathcal{G}, \alpha, \beta)$ from a family $\mathcal{Q}$~\cite{SJES979C35ACACA8723F5CDB6270E9D6E5E2}. The discrepancy between the two distributions is measured by the Kullback-Leibler (KL) divergence:
\begin{equation*}
	\mathrm{KL}(q \| p) = \int q(\mathcal{G}, \alpha, \beta) 
	\log \frac{q(\mathcal{G}, \alpha, \beta)}{p(\mathcal{G}, \alpha, \beta | y)} 
	\, d\mathcal{G} \, d\alpha \, d\beta
\end{equation*}
The goal of variational inference is to minimize this KL divergence. Using the mean-field approximation \cite{SJPD12102102372560}, we assume the variational distribution decomposes completely into the product of its factors:
\begin{equation*}
	q(\mathcal{G},\alpha,\beta)=q(\mathcal{G})q(\alpha)q(\beta)
\end{equation*}
Since the true posterior distribution $p(\mathcal{G},\alpha,\beta|y)$ is unknown, the KL divergence cannot be computed directly. To address this, we introduce the Evidence Lower Bound (ELBO):
\begin{equation*}
	\mathcal{L}(q)=\int q(\mathcal{G} ,\alpha,\beta)\log\frac{p(y,\mathcal{G},\beta,\alpha )}{q(\mathcal{G},\alpha,\beta)} d\mathcal{G}\ d\alpha \ d\beta 
\end{equation*}
The ELBO relates to the log marginal likelihood as follows:
\begin{equation*}
	\log p(y|\beta) =\mathcal{L}(q) +\mathrm{KL}(q|p)
\end{equation*}
Since the KL divergence is non-negative, the ELBO constitutes a lower bound on the log marginal likelihood. Therefore, maximizing the ELBO is equivalent to minimizing the KL divergence.
According to variational inference theory, the optimal variational factor satisfies:
\begin{equation*}
	\log q^*(z_i) = \mathbb{E}_{q_{-i}}[\log p(y, z)] + C
\end{equation*}
where $z = (\mathcal{G}, \alpha, \beta)$ denotes all latent variables, $q_{-i}$ 
represents the variational distributions of all variables except $z_i$, and 
$\mathbb{E}_{q_{-i}}[\cdot]$ denotes the expectation with respect to $q_{-i}$.

\subsubsection{Variational Distribution of the Core Tensor}

For the core tensor $\mathcal{G}$, when all other factors are fixed, the logarithm of its optimal variational distribution is:
\begin{equation*}
	\log q^*(\mathcal{G}) = \mathbb{E}_{q_{\alpha} q_{\beta}}
	[\log p(y|\mathcal{G}, \beta) + \log p(\mathcal{G}|\alpha)] + C
\end{equation*}
Substituting the observation likelihood $p(y|\mathcal{G}, \beta) = \mathcal{N}(y|\tilde{A}\boldsymbol{g}, \beta^{-1}I)$ 
and prior distribution $p(\mathcal{G}|\alpha) = \mathcal{N}(\boldsymbol{g}|0, \alpha^{-1}I)$ into the equation. Keeping only terms related to $\mathcal{G}$, we obtain:
\begin{equation*}
	\begin{split}
		\log q^*(\mathcal{G}) 
		&\propto -\frac{\mathbb{E}[\beta]}{2} \|y - \tilde{A}\boldsymbol{g}\|_{2}^{2} 
		-\frac{\mathbb{E}[\alpha]}{2} \|\boldsymbol{g}\|_{2}^{2} \\
		&= -\frac{1}{2} \boldsymbol{g}^\top 
		(\mathbb{E}[\beta]\tilde{A}^\top\tilde{A} + \mathbb{E}[\alpha]I)\boldsymbol{g} 
		+ \mathbb{E}[\beta]\boldsymbol{g}^\top\tilde{A}^\top y + C
	\end{split}
\end{equation*}
This expression is a quadratic form in $\boldsymbol{g}$. It indicates that $q^*(\mathcal{G})$ 
follow a Gaussian distribution:
\begin{equation}
	q^*(\mathcal{G}) = \mathcal{N}(\boldsymbol{g}|\mu_{\boldsymbol{g}}, \Sigma_{\boldsymbol{g}})
	\label{eq:23}
\end{equation}
Here, the posterior covariance and posterior mean are:
\begin{equation*}
	\Sigma_{\boldsymbol{g}} = (\mathbb{E}[\beta]\tilde{A}^\top\tilde{A} + \mathbb{E}[\alpha]I)^{-1}, 
	\quad 
	\mu_{\boldsymbol{g}} = \mathbb{E}[\beta]\Sigma_{\boldsymbol{g}}\tilde{A}^\top y
\end{equation*}
This result confirms that, even within the variational framework, the posterior distribution 
of the core tensor remains Gaussian. Its covariance matrix and posterior mean depend on the 
expected values of the hyperparameters $\mathbb{E}[\alpha]$ and $\mathbb{E}[\beta]$. Crucially, the 
posterior covariance matrix $\Sigma_{\boldsymbol{g}}$ has dimensions 
$n_{\mathcal{G}} \times n_{\mathcal{G}}$. This reduces the computational complexity of the matrix inversion  
from $O(n^3)$ to $O(n_{\mathcal{G}}^3)$. The result is a significant reduction in computational cost.

\subsubsection{Variational Distribution of Precision Hyperparameters}

For the signal precision parameter $\alpha$, the logarithmic form of its optimal variational distribution 
is:
\begin{equation*}
	\log q^*(\alpha) = \mathbb{E}_{q(\boldsymbol{g})}
	[\log p(\mathcal{G}|\alpha) + \log p(\alpha)] + C
\end{equation*}
Substituting the prior $p(\mathcal{G}|\alpha)$ and the hyperprior 
$p(\alpha) = \mathrm{Gamma}(\alpha|a_0, b_0)$, we get:
\begin{equation*}
	\begin{split}
		\log q^*(\alpha) 
		&\propto \frac{n_{\mathcal{G}}}{2}\log\alpha 
		- \frac{\alpha}{2}\mathbb{E}[\|\boldsymbol{g}\|_{2}^{2}] 
		+ (a_0 - 1)\log\alpha - b_0\alpha \\
		&= \left(a_0 + \frac{n_{\mathcal{G}}}{2} - 1\right)\log\alpha 
		- \left(b_0 + \frac{\mathbb{E}[\|\boldsymbol{g}\|_{2}^{2}]}{2}\right)\alpha
	\end{split}
\end{equation*}
where $n_{\mathcal{G}} = \prod_{k=1}^{d} R_k$ is the total number of elements in the 
core tensor. This expression shows $q^*(\alpha)$ is still a Gamma distribution:
\begin{equation}
	q^*(\alpha) = \mathrm{Gamma}(\alpha|\tilde{a}_{\alpha}, \tilde{b}_{\alpha})
	\label{eq:27}
\end{equation}
The update formulas for its shape and rate parameters are:
\begin{equation}
	\tilde{a}_{\alpha} = a_0 + \frac{n_{\mathcal{G}}}{2}, \quad 
	\tilde{b}_{\alpha} = b_0 + \frac{1}{2}\mathbb{E}_q[\|\boldsymbol{g}\|_{2}^{2}]
	\label{eq:28}
\end{equation}
For the Gaussian distribution $q(\mathcal{G}) = \mathcal{N}(\mu_{\boldsymbol{g}}, \Sigma_{\boldsymbol{g}})$, its second-order moment is
$\mathbb{E}_q[\|\boldsymbol{g}\|_{2}^{2}] = \|\mu_{\boldsymbol{g}}\|_{2}^{2} + \mathrm{tr}(\Sigma_{\boldsymbol{g}})$. 
Therefore, the expected precision is:
\begin{equation}
	\mathbb{E}[\alpha] = \frac{\tilde{a}_{\alpha}}{\tilde{b}_{\alpha}}
	\label{eq:29}
\end{equation}
Similarly, the optimal variational distribution for the noise precision $\beta$ is:
\begin{equation}
	q^*(\beta) = \mathrm{Gamma}(\beta|\tilde{a}_{\beta}, \tilde{b}_{\beta})
	\label{eq:30}
\end{equation}
with parameters:
\begin{equation}
	\tilde{a}_{\beta} = a_0 + \frac{m}{2}, \quad 
	\tilde{b}_{\beta} = b_0 + \frac{1}{2}\mathbb{E}_q[\|y - \tilde{A}\boldsymbol{g}\|_{2}^{2}]
	\label{eq:31}
\end{equation}
The posterior estimate of the noise variance is:
\begin{equation}
	\hat{\sigma}^2 = \frac{1}{\mathbb{E}[\beta]} = \frac{\tilde{b}_{\beta}}{\tilde{a}_{\beta}} 
	= \frac{b_0 + \frac{1}{2}\left(\|y - \tilde{A}\mu_{\boldsymbol{g}}\|_{2}^{2} 
		+ \mathrm{tr}(\tilde{A}^\top\tilde{A}\Sigma_{\boldsymbol{g}})\right)}{a_0 + \frac{m}{2}}
	\label{eq:32}
\end{equation}
Under weak prior information ($a_0, b_0 \to 0$), this simplifies $\hat{\sigma}^2$ approximates:
\begin{equation*}
	\hat{\sigma}^2 \approx \frac{\|y - \tilde{A}\mu_{\boldsymbol{g}}\|_{2}^{2} 
		+ \mathrm{tr}(\tilde{A}^\top\tilde{A}\Sigma_{\boldsymbol{g}})}{m}
\end{equation*}
The regularization parameter $\lambda$ is expressed as the ratio of the two expected 
precision parameters:
\begin{equation}
	\lambda = \frac{\mathbb{E}[\alpha]}{\mathbb{E}[\beta]} 
	= \frac{\tilde{a}_{\alpha} \tilde{b}_{\beta}}{\tilde{b}_{\alpha} \tilde{a}_{\beta}}
	\label{eq:lambda}
\end{equation}
Based on the derivations above, we propose the Tucker-VB algorithm. Algorithm~1 
describes the variation for a single hyperparameter. It uses a single precision 
parameter $\alpha$ to impose a global regularization constraint on the core tensor. 
This algorithm outputs the estimated regularization parameter $\hat{\lambda}$ and 
noise variance $\hat{\sigma}^2$.

\begin{algorithm}[htbp]
	\caption{Tucker-VB with single hyperparameter}
	\label{alg:1}
	\KwIn{ $y, \tilde{A}, a_0, b_0, \varepsilon, T_{\max}$ }
	\KwOut{ $\hat{\lambda}, \hat{\sigma}^2$ }
	Initialize $\mathbb{E}[\alpha] \leftarrow 1$, $\mathbb{E}[\beta] \leftarrow 1$\;
	Precompute $\tilde{A}^T \tilde{A}$, $\tilde{A}^T y$\;
	\For{$t = 1, 2, \dots, T_{\max}$}{
		Compute $\Sigma_g$ via Eq. \eqref{eq:23}\;
		Compute $\mu_g$ via Eq. \eqref{eq:23}\;
		Compute $\mathbb{E}\left[\|\mathbf{g}\|^2\right] = \|\mu_g\|^2 + \text{tr}(\Sigma_g)$\;
		Update $\tilde{a}_\alpha, \tilde{b}_\alpha$ via Eq. \eqref{eq:28}\;
		Compute $\mathbb{E}[\alpha] = \tilde{a}_\alpha / \tilde{b}_\alpha$\;
		Compute $\mathbb{E}\left[\|y - \tilde{A}\mathbf{g}\|^2\right] = \|y - \tilde{A}\mu_g\|^2 + \text{tr}\left(\tilde{A}^T \tilde{A}\Sigma_g\right)$\;
		Update $\tilde{a}_\beta, \tilde{b}_\beta$ via Eq. \eqref{eq:31}\;
		Compute $\mathbb{E}[\beta] = \tilde{a}_\beta / \tilde{b}_\beta$\;
		\If{convergence criterion met}{
			{\bf Break}
		}
	}
	Compute $\hat{\lambda}$ via Eq. \eqref{eq:lambda}\;
	Compute $\hat{\sigma}^2$ via Eq. \eqref{eq:32}\;
\end{algorithm}

\subsection{Anisotropic Multi-Hyperparameter Prior Structure}

The single-hyperparameter model in Section 3.2 assigns a uniform precision parameter $\alpha$ to all elements of the core tensor $\mathcal{G}$. This imposes isotropic regularization. However, this approach is often insufficient for many practical problems. A key issue arises when different tensor modes possess fundamentally different physical characteristics. In such cases, they require differentiated regularization strengths. Consider dynamic medical image reconstruction as an example. The unknown quantity is represented as a 3D tensor $\mathcal{X} \in R^{I_1 \times I_2 \times I_3}$ with two spatial modes and one temporal mode. The temporal mode captures rapidly changing dynamics. It typically requires weak regularization to preserve fine temporal details. In contrast, the spatial modes contain slowly varying anatomical structures. These are well-suited for stronger regularization to suppress noise. Similarly, in anisotropic thermal conduction inverse problems, signals decay at different rates along different axes due to direction-dependent thermal diffusivity. This necessitates a direction-specific regularization strategy. A single regularization parameter cannot flexibly handle such anisotropic properties. To achieve differentiated regularization across tensor modes, we leverage the multilinear structure of the Tucker decomposition. We extend the prior distribution to a multi-hyperparameter form. We assign independent precision hyperparameters to different modes of the core tensor $\mathcal{G}$. This enables mode-adaptive regularization.

The core tensor $\mathcal{G} \in R^{R_1 \times R_2 \times \cdots \times R_d}$ possesses a $d$th-order multilinear structure. We define the $i_k$-th slice vector in mode $k$ as $\boldsymbol{g}_{i_k}^{(k)} \in R^{n_{\mathcal{G}}/R_k}$ (where $i_k = 1, \ldots, R_k$). This vector contains the vectorized elements where the $k$-th index fixed at $i_k$. To control the regularization strength of each mode, we introduce $d$ independent precision hyperparameters 
$\alpha_1, \alpha_2, \ldots, \alpha_d$. A conditionally independent Gaussian prior distribution is assigned to the slice vectors of each mode:
\begin{equation*}
	p(\boldsymbol{g}_{i_k}^{(k)} | \alpha_k) = 
	\mathcal{N}(\boldsymbol{g}_{i_k}^{(k)} | 0, \alpha_k^{-1} I)
\end{equation*}
The joint prior distribution can be expressed as:
\begin{equation*}
	p(\mathcal{G} | \alpha_1, \alpha_2, \ldots, \alpha_d) \propto 
	\exp\left\{-\frac{1}{2} \sum_{k=1}^{d} \alpha_k 
	\sum_{i_k=1}^{R_k} \|\boldsymbol{g}_{i_k}^{(k)}\|^2 \right\}
\end{equation*}
A key observation is that for any mode $k$, the sum of squared norms across all slices 
equals the total squared Frobenius norm of the core tensor:
\begin{equation*}
	\sum_{i_k=1}^{R_k} \|\boldsymbol{g}_{i_k}^{(k)}\|^2 = \|\boldsymbol{g}\|^2, 
	\quad \forall k = 1, \ldots, d
\end{equation*}
Consequently, the joint prior distribution simplifies to:
\begin{equation*}
	p(\mathcal{G} | \alpha_1, \alpha_2, \ldots, \alpha_d) \propto 
	\exp\left\{-\frac{1}{2} \left(\sum_{k=1}^{d} \alpha_k\right) \|\boldsymbol{g}\|_2^2 \right\} 
	= \exp\left(-\frac{\bar{\alpha}}{2} \|\boldsymbol{g}\|_2^2\right)
\end{equation*}
where $\bar{\alpha} = \sum_{k=1}^{d} \mathbb{E}[\alpha_k]$ is the effective precision parameter. This formulation reveals that, the multi-mode prior is formally equivalent to an isotropic Gaussian prior on the entire core tensor with effective precision $\bar{\alpha}$. The critical distinction, however, is that $\bar{\alpha}$ comprises $d$ independently evolving components $\{\alpha_k\}_{k=1}^{d}$. During variational inference, each $\alpha_k$ is updated independently based on the data-model fit for its corresponding mode. This adaptation to mode-specific regularization requirements enables anisotropic regularization.

Each precision parameter is independently assigned a Gamma prior:
\begin{equation*}
	p(\alpha_k) = \mathrm{Gamma}(\alpha_k | a_0, b_0), \quad k = 1, \ldots, d
\end{equation*}
Under the mean-field variational approximation, the posterior distribution is assumed to factorize as:
\begin{equation*}
	q(\mathcal{G}, \{\alpha_k\}_{k=1}^{d}, \beta) = 
	q(\mathcal{G}) \prod_{k=1}^{d} q(\alpha_k) \cdot q(\beta)
\end{equation*}
The optimal variational distribution for the core tensor remains Gaussian, 
$q^*(\boldsymbol{g}) = \mathcal{N}(\boldsymbol{g} | \mu_{\boldsymbol{g}}, \Sigma_{\boldsymbol{g}})$, 
with:
\begin{equation*}
	\Sigma_{\boldsymbol{g}} = \left(\mathbb{E}[\beta] \tilde{A}^\top \tilde{A} + \bar{\alpha} I\right)^{-1}, 
	\quad 
	\mu_{\boldsymbol{g}} = \mathbb{E}[\beta] \Sigma_{\boldsymbol{g}} \tilde{A}^\top y
\end{equation*}
For the precision parameter $\alpha_k$ of mode $k$, the optimal variational distribution satisfies:
\begin{equation*}
	\log q^*(\alpha_k) = \mathbb{E}_{\mathcal{G}, \{\alpha_{j \neq k}\}}
	[\log p(\mathcal{G} | \boldsymbol{\alpha}) + \log p(\alpha_k)] + \mathrm{const}
\end{equation*}
where $\boldsymbol{\alpha} = (\alpha_1, \ldots, \alpha_d)^\top$ is the precision parameter vector. Substituting the prior distributions and expanding yields:
\begin{equation}
	\begin{split}
		\log q^*(\alpha_k) 
		&\propto \frac{R_k \cdot (n_{\mathcal{G}} / R_k)}{2} \log \alpha_k 
		- \frac{\alpha_k}{2} \mathbb{E}[\|\boldsymbol{g}\|^2] 
		+ (a_0 - 1) \log \alpha_k - b_0 \alpha_k \\
		&= \left(a_0 + \frac{n_{\mathcal{G}}}{2} - 1\right) \log \alpha_k 
		- \left(b_0 + \frac{\mathbb{E}[\|\boldsymbol{g}\|^2]}{2}\right) \alpha_k
	\end{split}
	\label{eq:43}
\end{equation}
Therefore, the optimal variational distribution for the $k$-th mode precision parameter remains a Gamma distribution as in Eq.~\eqref{eq:44}:
\begin{equation}
	q^*(\alpha_k) = \mathrm{Gamma}(\alpha_k | \tilde{a}_k, \tilde{b}_k)
	\label{eq:44}
\end{equation}
The updated parameters are:
\begin{equation}
	\tilde{a}_k = a_0 + \frac{n_{\mathcal{G}}}{2}, \quad 
	\tilde{b}_k = b_0 + \frac{1}{2} \mathbb{E}[\|\boldsymbol{g}\|^2]
	\label{eq:45}
\end{equation}
The expected value is $\mathbb{E}[\alpha_k] = \tilde{a}_k / \tilde{b}_k$. The 
regularization parameter for mode $k$ is defined as:
\begin{equation}
	\lambda_k = \hat{\sigma}^2 \mathbb{E}[\alpha_k] = 
	\hat{\sigma}^2 \frac{\tilde{a}_k}{\tilde{b}_k}, \quad k = 1,\ldots, d
    \label{eq:lambdak}
\end{equation}
These mode-specific regularization parameters $\lambda_k$ independently control the regularization strength for each mode of the core tensor.

The core advantage of the multi-hyperparameter model lies in its ability to automatically learn the optimal regularization strength for each mode from observed data. This adaptive process is entirely data-driven, requiring neither manual tuning nor domain expertise. The algorithm automatically adjusts mode-specific regularization strengths based on the degree of fit between observations and model assumptions. When a particular pattern exhibits high information content and rapid variation, the algorithm tends to select smaller accuracy parameters, reducing regularization to preserve information. Conversely, when a pattern is smoother with significant noise influence, the algorithm automatically increases accuracy parameters to strengthen regularization. This mode-wise adaptive regularization mechanism enables flexible adaptation to diverse data characteristics, yielding significant performance gains for inverse problems with complex anisotropic structures. Algorithm~2 give the complete implementation of the multi-hyperparameter Tucker-VB algorithm. By introducing independent precision hyperparameters $\{\alpha_k\}_{k=1}^d$ for the $d$ modes of the core tensor, the algorithm achieves anisotropic differentiated regularization. The outputs include mode-specific regularization parameters $\{\hat{\lambda}_k\}_{k=1}^d$ and the estimated noise variance $\hat{\sigma}^2$.

\begin{algorithm}[htbp]
	\caption{Tucker-VB with multiple hyperparameters}
	\label{alg:2}
	\KwIn{ $y, \tilde{A}, d, a_0, b_0, \varepsilon, T_{\max}$ }
	\KwOut{ $\{\hat{\lambda}_k\}_{k=1}^d, \hat{\sigma}^2$ }
	Initialize $\mathbb{E}[\alpha_k] \leftarrow 1$ for $k = 1, \dots, d$; $\mathbb{E}[\beta] \leftarrow 1$\;
	Precompute $\tilde{A}^T \tilde{A}$, $\tilde{A}^T y$, $n_G = \prod_{k=1}^d R_k$\;
	\For{$t = 1, 2, \dots, T_{\max}$}{
		Compute $\bar{\alpha} = \sum_{k=1}^d \mathbb{E}[\alpha_k]$\;
		Compute $\Sigma_g = \left(\mathbb{E}[\beta] \tilde{A}^T \tilde{A} + \bar{\alpha} I\right)^{-1}$\;
		Compute $\mu_g = \mathbb{E}[\beta] \Sigma_g \tilde{A}^T y$\;
		Compute $\mathbb{E}\left[\|\mathbf{g}\|^2\right] = \|\mu_g\|^2 + \text{tr}(\Sigma_g)$\;
		\For{$k = 1, \dots, d$}{
			Update $\tilde{a}_k, \tilde{b}_k$ via Eq. \eqref{eq:45}\;
			Compute $\mathbb{E}[\alpha_k] = \tilde{a}_k / \tilde{b}_k$\;
		}
		Compute $\mathbb{E}\left[\|y - \tilde{A}\mathbf{g}\|^2\right] = \|y - \tilde{A}\mu_g\|^2 + \text{tr}\left(\tilde{A}^T \tilde{A}\Sigma_g\right)$\;
		Update $\tilde{a}_\beta = a_0 + \frac{m}{2}$, $\tilde{b}_\beta = b_0 + \frac{1}{2}\mathbb{E}\left[\|y - \tilde{A}\mathbf{g}\|^2\right]$\;
		Compute $\mathbb{E}[\beta] = \tilde{a}_\beta / \tilde{b}_\beta$\;
		\If{convergence criterion met}{
			{\bf Break}
		}
	}
	Compute $\hat{\sigma}^2 = 1/\mathbb{E}[\beta]$\;
	Compute $\hat{\lambda}_k$ for $k = 1, \dots, d$ via Eq. \eqref{eq:lambdak}\;
\end{algorithm}

\section{Numerical Experiments}
\subsection{Algorithm Validation}

The Tucker-VB method is validated here through systematic comparison against direct Variational Bayes (VB). Two aspects receive attention: (i) reconstruction accuracy and noise estimation under varying noise levels; (ii) how problem dimensionality affects computational efficiency and scalability. A two-dimensional separable integral equation~\cite{SJES2C4712DCFEDF50DA7FB514E6E2A52B68} provides the test case, with the forward operator exhibiting Kronecker product structure. The true solution derives from outer products of low-frequency sinusoidal functions, and observations come from applying the forward operator with additive Gaussian white noise. Four metrics are tracked: relative reconstruction error, noise estimation error, computation time, and speedup ratio (the latter quantifying efficiency gains from Tucker low-rank constraints). Two experimental groups probe the effects of noise level and problem dimensionality. All results represent averages over 100 independent trials; means and standard deviations are reported throughout.

Table~\ref{tab:reconstruction_error} shows reconstruction error for both methods across noise levels. The relative error gap between Tucker-VB and direct VB shrinks monotonically from 27.6\% to 8.1\% as noise level $\sigma$ rises from 0.01 to 0.20, noise level clearly governs the relative performance of these two approaches.
\begin{table*}[!ht]
	\centering
	\caption{Reconstruction error comparison across noise levels}
	\begin{tabular}{l c c c}
		\toprule
		Noise Level $\sigma$ & Tucker-VB & VB & Relative Gap \\ \midrule
		0.01 & $0.0319 \pm 0.0007$ & $0.0250 \pm 0.0006$ & $+27.6\%$ \\
		0.02 & $0.0420 \pm 0.0022$ & $0.0349 \pm 0.0017$ & $+20.3\%$ \\
		0.05 & $0.0830 \pm 0.0069$ & $0.0730 \pm 0.0050$ & $+13.7\%$ \\
		0.10 & $0.1548 \pm 0.0138$ & $0.1391 \pm 0.0101$ & $+11.3\%$ \\
		0.20 & $0.2809 \pm 0.0244$ & $0.2598 \pm 0.0192$ & $+8.1\%$  \\
		\bottomrule
	\end{tabular}
	\label{tab:reconstruction_error}
\end{table*}

Tucker-VB shows higher error than direct VB at $\sigma = 0.01$; at $\sigma = 0.20$, this gap has narrowed substantially. Low-noise observations carry rich high-frequency information, which direct VB extracts effectively through its complete singular value structure. Tucker-VB's low-rank approximation truncates small singular values and loses fine details as a result. When noise increases, high-frequency components become noise-dominated, and excessive singular values introduce contamination rather than useful signal. Tucker-VB's rank constraint then acts as implicit regularization, narrowing the performance gap.

Table~\ref{tab:noise_estimation} reports noise estimation results: automatic noise parameter estimation being a principal advantage of variational Bayesian methods over classical regularization. Both Tucker-VB and direct VB estimate the noise level with high accuracy across the tested range.

\begin{table*}[!ht]
	\centering
	\caption{Noise estimation accuracy across noise levels}
	\begin{tabular}{l c c c c}
		\toprule
		Noise Level $\sigma$ & Tucker-VB $\hat{\sigma}$ & Rel. Error & VB $\hat{\sigma}$ & Rel. Error \\ \midrule
		0.01 & $0.0103 \pm 0.0001$ & $+3.0\%$ & $0.0101 \pm 0.0001$ & $+1.0\%$ \\
		0.02 & $0.0201 \pm 0.0002$ & $+0.5\%$ & $0.0200 \pm 0.0002$ & $0.0\%$  \\
		0.05 & $0.0499 \pm 0.0006$ & $-0.2\%$ & $0.0499 \pm 0.0006$ & $-0.2\%$ \\
		0.10 & $0.0997 \pm 0.0012$ & $-0.3\%$ & $0.0997 \pm 0.0012$ & $-0.3\%$ \\
		0.20 & $0.1992 \pm 0.0024$ & $-0.4\%$ & $0.1993 \pm 0.0024$ & $-0.4\%$ \\
		\bottomrule
	\end{tabular}
	\label{tab:noise_estimation}
\end{table*}
Low-rank approximation does not degrade Tucker-VB's noise estimates. Average relative errors are 0.6\% for Tucker-VB and 0.4\% for direct VB, a negligible difference. Signal reconstruction suffers somewhat under the low-rank constraint, but residual statistics remain similar between the two methods, so noise estimates coincide. At $\sigma \geq 0.05$, both methods return identical estimates to four decimal places.

Table~\ref{tab:computation_time} compares computation time. Tucker-VB runs far faster on large problems; more importantly, its scaling behavior with problem size differs qualitatively from that of direct VB.
\begin{table*}[!ht]
	\centering
	\caption{Computation time comparison across problem dimensions}
	\begin{tabular}{l c c c c}
		\toprule
		Dimension $n$ & Problem Size $n^2$ & Tucker-VB & VB & Speedup \\ 
		\midrule
		30  & 900    & $0.0012 \pm 0.0031$ s & $0.109 \pm 0.020$ s & $91\times$     \\
		50  & 2,500  & $0.0016 \pm 0.0030$ s & $5.70 \pm 0.38$ s   & $3,563\times$  \\
		80  & 6,400  & $0.0027 \pm 0.0031$ s & $111.6 \pm 7.6$ s   & $41,333\times$ \\
		100 & 10,000 & $0.0019 \pm 0.0037$ s & N/A                 & ---            \\
		150 & 22,500 & $0.0050 \pm 0.0056$ s & N/A                 & ---            \\
		200 & 40,000 & $0.0085 \pm 0.0045$ s & N/A                 & ---            \\
		\bottomrule
	\end{tabular}
	\label{tab:computation_time}
\end{table*}
Direct VB constructs the full $n^2 \times n^2$ system matrix and computes its SVD, giving $O(n^6)$ complexity. Tucker-VB leverages the operator's low-rank structure and runs in approximately $O(n^2)$ time, four orders of magnitude less.
The timing data for direct VB illustrate this scaling: 0.109 seconds at $n=30$, 5.70 seconds at $n=50$, and 111.6 seconds at $n=80$, corresponding to roughly 52-fold and 20-fold increases. Such super-linear growth fits the $O(n^6)$ complexity profile. Tucker-VB timings tell a different story: 1.2 milliseconds at $n=30$ climbing to just 8.5 milliseconds at $n=200$. A nearly 7-fold increase in dimension produces only a 6-fold increase in computation time, consistent with near-linear scaling.
The ``N/A'' entries expose a hard limitation of direct VB: memory exhaustion beyond $n=80$. At $n=200$, storage requirements reach roughly 128GB, well beyond typical computing resources. Tucker-VB requires only $O(n^2 + r^2)$ memory, several megabytes at $n=200$. These experiments establish Tucker-VB as the preferred choice for medium to large-scale inverse problems.

\subsection{2D Fredholm Integral Equation of the First Kind}

We consider the inverse problem for two-dimensional Fredholm integral equations of the first kind~\cite{SSJD121114000286}. Let $\Omega = [0,1]^2$; the forward problem reads:
\begin{equation*}
	g(s,t) = \int_{0}^{1} \int_{0}^{1} K(s,x) K(t,y) f(x,y) \, dx \, dy, 
	\quad (s,t) \in \Omega
\end{equation*}
where $f(x,y)$ is the unknown source function, $g(s,t)$ is the response function, and $K(\cdot , \cdot)$ is the integral kernel. The kernel admits a spectral decomposition:
\begin{equation*}
	K(s,x) = \sum_{k=1}^{n} \sigma_k \varphi_k(s) \varphi_k(x)
\end{equation*}
with sinusoidal basis functions $\varphi_k(x) = \sqrt{2} \sin(k\pi x)$ satisfying homogeneous Dirichlet boundary conditions. The singular values $\sigma_k = e^{-\alpha k}$ decay at a rate governed by $\alpha > 0$, yielding condition number $\kappa = e^{\alpha(n-1)}$ for the one-dimensional operator and $\kappa_{2D} = \kappa^2$ for the two-dimensional problem. Only noisy observations $g^\delta(s,t)$ are available ($\delta > 0$ denoting noise level). The goal is to recover $f$ from $g^\delta$. The true solution takes the form of a multi-peak Gaussian:
\begin{equation*}
	f(x,y) = \sum_{j=1}^{5} a_j \exp\left(-\frac{(x - c_j^x)^2 + (y - c_j^y)^2}{2\sigma_j^2}\right)
\end{equation*}
Experimental setup: grid size $n = 150$ (22,500 unknowns), kernel decay rate $\alpha = 0.15$, noise levels $\delta \in \{0.001, 0.002, 0.005, 0.01, 0.02, 0.05, 0.1\}$. Each noise level runs for 500 independent trials. Four classical regularization parameter selection methods serve as baselines. Relative reconstruction error is the evaluation metric:
\begin{equation*}
	e_{\mathrm{rel}} = \frac{\|\hat{F} - F\|_F}{\|F\|_F}
\end{equation*}
where $\|\cdot\|_F$ denotes the Frobenius norm.

\begin{figure*}
	\centering
	\includegraphics[width=\textwidth]{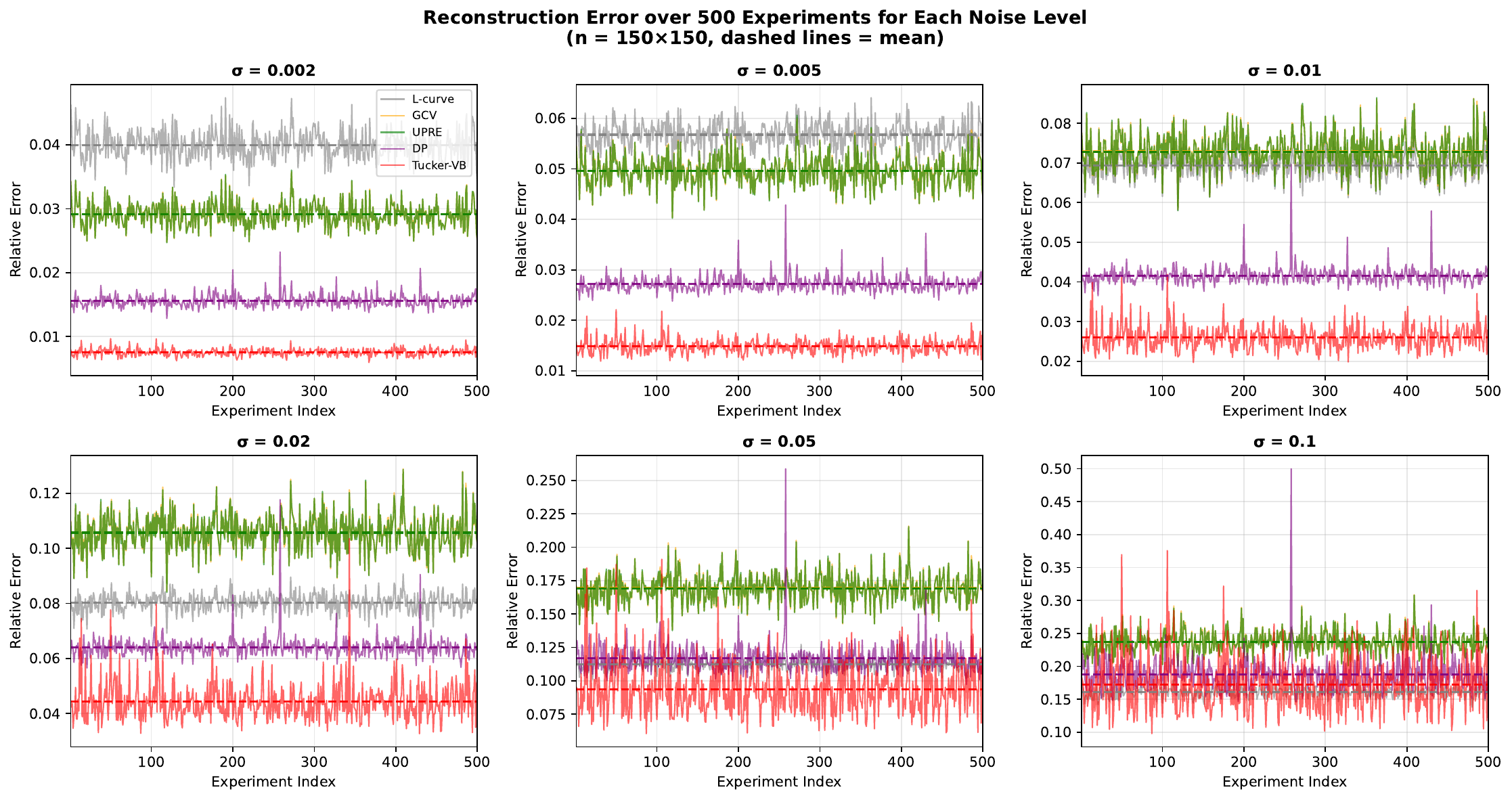}
	\caption{Reconstruction error curves from 500 independent trials at six noise 
		levels: L-curve (gray), GCV (orange), UPRE (green), DP (purple), and Tucker-VB (red).}
	\label{fig:error_curves}
\end{figure*}

Figure~\ref{fig:error_curves} plots reconstruction error curves from 500 trials. Tucker-VB (red) achieves the lowest error at all six noise levels. At moderate noise ($\sigma = 0.02$), Tucker-VB averages roughly 0.04, versus 0.065 for DP and 0.10 for UPRE.
Tucker-VB also shows the smallest fluctuation amplitude, indicating stable performance across noise realizations. UPRE and DP, it should be noted, use the true noise variance in these experiments---information rarely available in practice. Tucker-VB, fully automatic, still outperforms both.
The multi-parameter adaptive mechanism underlies this behavior: independent regularization parameters for different Tucker modes allow the algorithm to identify dominant signal components and penalize them lightly while suppressing noise-dominated components more aggressively.

\begin{figure*}
	\centering
	\includegraphics[width=\textwidth]{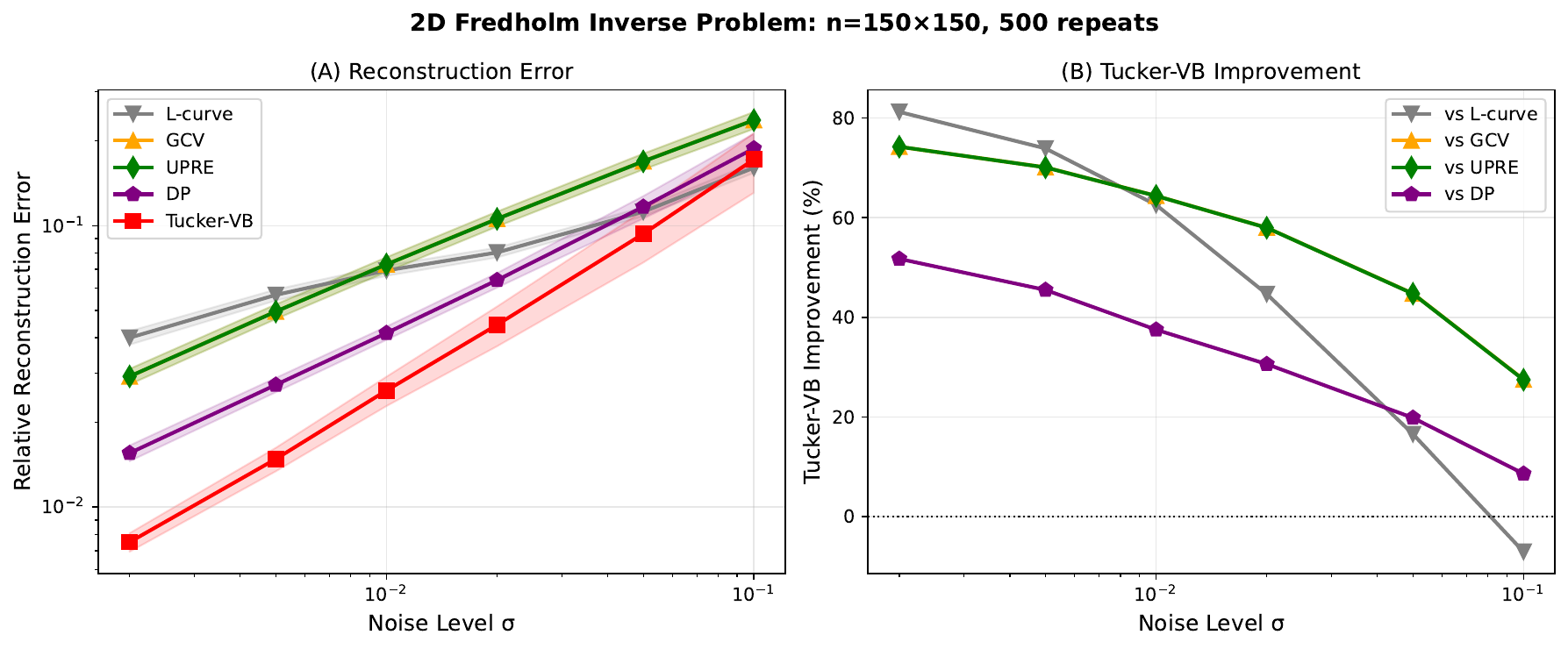}
	\caption{(A) Relative reconstruction error versus noise level $\sigma$ for five 
		regularization methods; shaded regions indicate $\pm 1$ standard deviation. 
		(B) Percentage improvement of Tucker-VB over other methods.}
	\label{fig:fredholm_results}
\end{figure*}

Figure~\ref{fig:fredholm_results}(A) shows reconstruction error versus noise level. Tucker-VB (red) achieves the lowest error throughout, and the gap persists across the entire noise range. All methods see error grow with noise, but Tucker-VB grows more slowly.
Figure~\ref{fig:fredholm_results}(B) quantifies the percentage improvement of Tucker-VB over each baseline. Improvement shrinks as noise rises, yet Tucker-VB retains roughly 20\% gain even in the high-noise regime. At $\sigma = 0.1$, improvement over DP drops to near zero or slightly negative (around $-5\%$). DP becomes competitive under extreme noise because it exploits prior knowledge of the noise level---a luxury seldom available in practice.
Tucker-VB yields 20\%--50\% average improvement over traditional methods on this 2D Fredholm problem. It attains top reconstruction accuracy across a broad noise range, estimates the unknown noise level reliably, and tunes regularization strength per directional component, balancing noise suppression against signal fidelity.

\subsection{Image Deblurring}

Two-dimensional image deblurring experiments test algorithm performance on the $512 \times 512$ Cameraman image, with pixel values normalized to $[0,1]$. This image contains edges, textures, and smooth regions suitable for evaluating deblurring methods. Two noise levels are tested: $\sigma = 0.01$ and $\sigma = 0.03$.
Two standard image quality metrics are used. Peak Signal-to-Noise Ratio (PSNR), based on mean squared error:
\begin{equation*}
	\mathrm{PSNR} = 10 \log_{10}\left(\frac{\mathrm{MAX}^2}{\mathrm{MSE}}\right) 
	= 10 \log_{10}\left(\frac{\mathrm{MAX}^2}{\frac{1}{n^2}\|\hat{X} - X\|_F^2}\right)
\end{equation*}
where $\mathrm{MAX}$ is the maximum pixel value ($\mathrm{MAX} = 1$ here), $\hat{X}$ is the reconstructed image, and $X$ is the original.
Structural Similarity Index (SSIM) measures luminance, contrast, and structure:
\begin{equation*}
	\mathrm{SSIM}(X, \hat{X}) = 
	\frac{(2\mu_X \mu_{\hat{X}} + C_1)(2\sigma_{X\hat{X}} + C_2)}
	{(\mu_X^2 + \mu_{\hat{X}}^2 + C_1)(\sigma_X^2 + \sigma_{\hat{X}}^2 + C_2)}
\end{equation*}
where $\mu_X$, $\mu_{\hat{X}}$ are image means, $\sigma_X^2$, $\sigma_{\hat{X}}^2$ are variances, and $\sigma_{X\hat{X}}$ is covariance. Stabilization constants are $C_1 = (K_1 \cdot \mathrm{MAX})^2$ and $C_2 = (K_2 \cdot \mathrm{MAX})^2$, with $K_1 = 0.01$, $K_2 = 0.03$.

Table~\ref{tab:comprehensive_comparison} lists PSNR, SSIM, and Tucker-VB's improvement over other methods. Tucker-VB achieves the best reconstruction at both noise levels: PSNR 30.27~dB and SSIM 0.9941 at $\sigma = 0.01$; PSNR 27.55~dB and SSIM 0.9889 at $\sigma = 0.03$.
\begin{table*}[!ht]
	\centering
	\caption{Reconstruction performance comparison across noise levels}
	\begin{tabular}{l l c c c c}
		\toprule
		Noise Level & Method & PSNR (dB) & SSIM & $\Delta$PSNR (dB) & $\Delta$SSIM \\ \midrule
		$\sigma=0.01$ & L-curve   & 17.80 & 0.9022 & +12.47 & +0.0919 \\
		& GCV       & 26.82 & 0.9870 & +3.45  & +0.0071 \\
		& UPRE      & 26.66 & 0.9865 & +3.61  & +0.0076 \\
		& DP        & 29.54 & 0.9931 & +0.73  & +0.0010 \\
		& Tucker-VB & \textbf{30.27} & \textbf{0.9941} & --- & --- \\ \midrule
		$\sigma=0.03$ & L-curve   & 16.87 & 0.8800 & +10.68 & +0.1089 \\
		& GCV       & 22.39 & 0.9642 & +5.16  & +0.0247 \\
		& UPRE      & 22.47 & 0.9649 & +5.08  & +0.0240 \\
		& DP        & 25.46 & 0.9826 & +2.09  & +0.0063 \\
		& Tucker-VB & \textbf{27.55} & \textbf{0.9889} & --- & --- \\
		\bottomrule
	\end{tabular}
	\label{tab:comprehensive_comparison}
\end{table*}
The ranking holds at both noise levels: Tucker-VB $>$ DP $>$ GCV $\approx$ UPRE $>$ L-curve.
Tucker-VB's largest gains come against L-curve: +12.47~dB PSNR and +0.0919 SSIM at $\sigma = 0.01$; +10.68~dB and +0.1089 at $\sigma = 0.03$. L-curve selects regularization parameters from the curvature of the residual-norm versus solution-norm plot, a heuristic that tends to over-regularize ill-conditioned problems and blur out details.
Tucker-VB also beats GCV and UPRE by clear margins: +3.45~dB and +3.61~dB at $\sigma = 0.01$, expanding to +5.16~dB and +5.08~dB at $\sigma = 0.03$.
The gap over DP is smaller but still meaningful: +0.73~dB at $\sigma = 0.01$, +2.09~dB at $\sigma = 0.03$. DP uses exact noise level information, while Tucker-VB estimates it from the data alone---yet Tucker-VB still wins.

\begin{figure*}
	\centering
	\includegraphics[width=\textwidth]{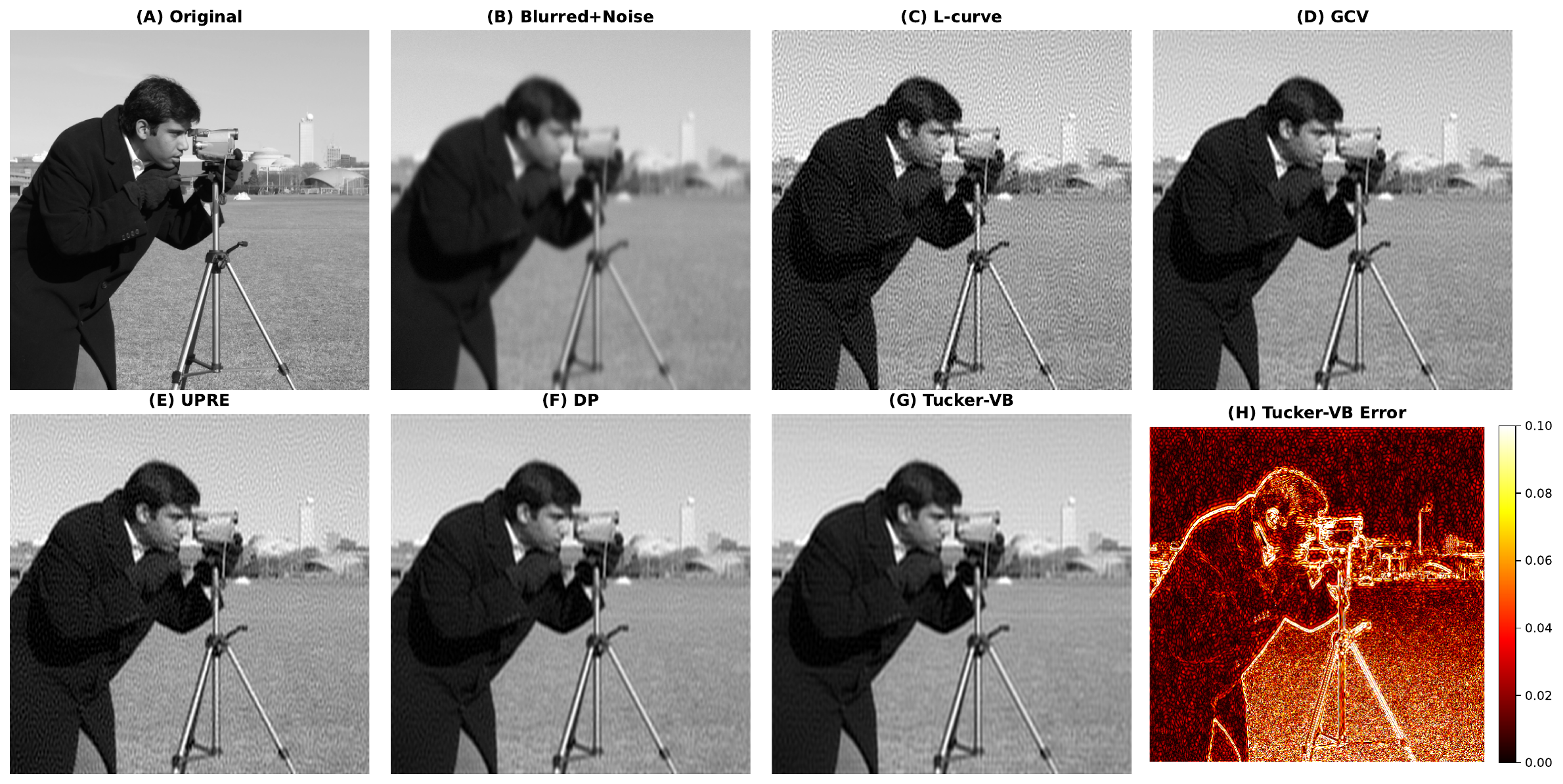}
	\caption{Anisotropic Gaussian deblurring results ($\sigma = 0.01$). 
		(A) Original image; (B) Blurred and noisy observation; (C) L-curve reconstruction; 
		(D) GCV reconstruction; (E) UPRE reconstruction; (F) DP reconstruction; 
		(G) Tucker-VB reconstruction; (H) Tucker-VB absolute error map.}
	\label{fig:deblur_001}
\end{figure*}

\begin{figure*}
	\centering
	\includegraphics[width=\textwidth]{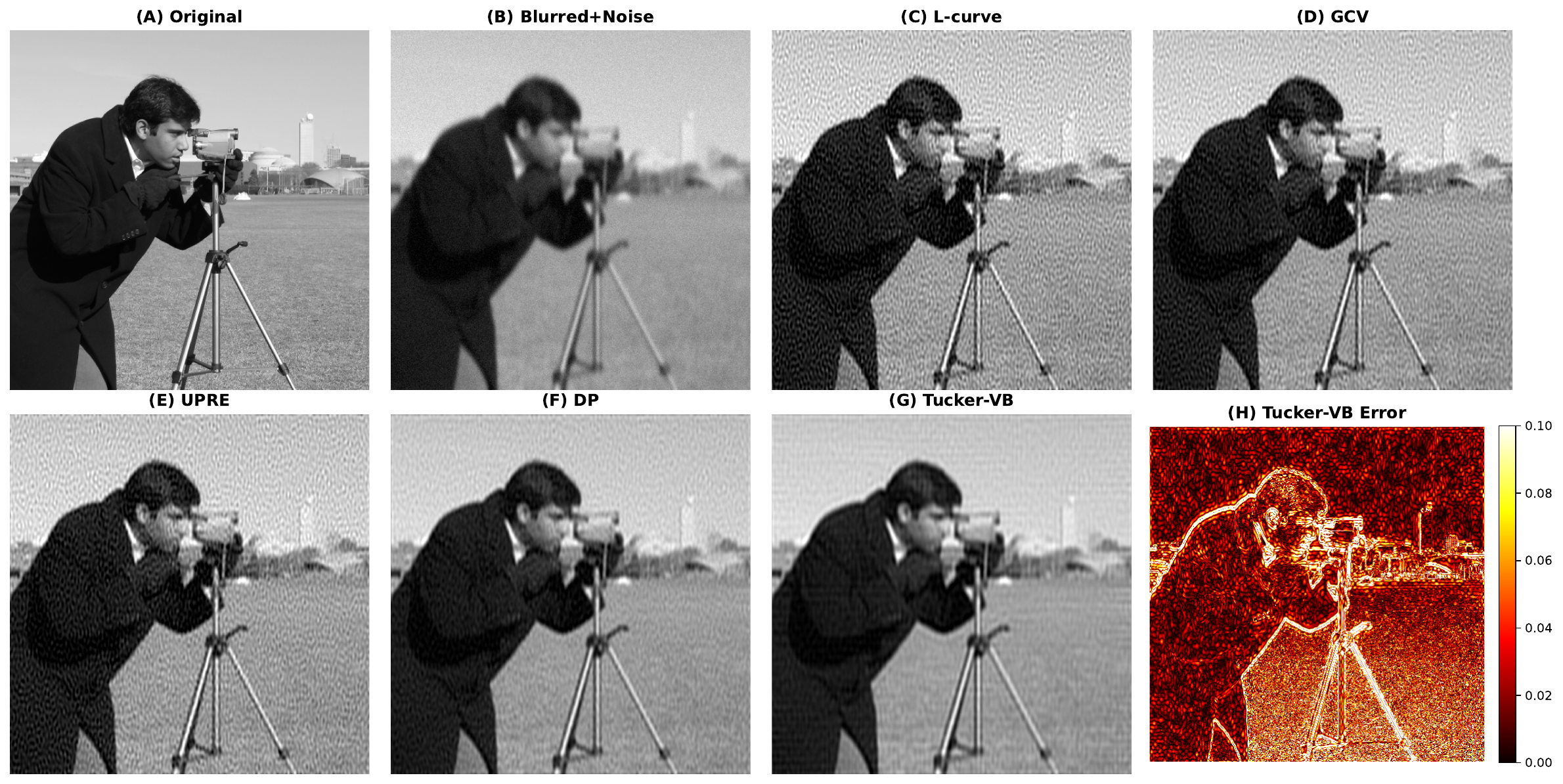}
	\caption{Anisotropic Gaussian deblurring results ($\sigma = 0.03$). 
		(A) Original image; (B) Blurred and noisy observation; (C) L-curve reconstruction; 
		(D) GCV reconstruction; (E) UPRE reconstruction; (F) DP reconstruction; 
		(G) Tucker-VB reconstruction; (H) Tucker-VB absolute error map.}
	\label{fig:deblur_003}
\end{figure*}

Figures~\ref{fig:deblur_001} and \ref{fig:deblur_003} show reconstructions at both noise levels.

L-curve (panel C) over-smooths severely; details vanish and only coarse contours survive, consistent with its low PSNR (17.80~dB and 16.87~dB). GCV (panel D) leaves visible residual noise, especially at $\sigma = 0.03$, single-parameter regularization cannot balance noise suppression and detail preservation well. DP (panel E) outperforms GCV with clearer output and less noise, though some blurring persists. Tucker-VB (panel F) produces the sharpest edges and best-preserved textures, closest to the original.
The error heatmap (panel H) shows absolute error $|f_{\mathrm{true}} - \hat{f}|$ for Tucker-VB. Errors concentrate at edges (human silhouette, tripod), while smooth regions (sky, grass) have smaller errors. Edge high-frequency content suffers the most attenuation during blurring and is hardest to recover. Comparing heatmaps across noise levels, errors at $\sigma = 0.03$ are larger overall, but the spatial pattern stays similar.

\begin{figure*}
	\centering
	\includegraphics[width=\textwidth]{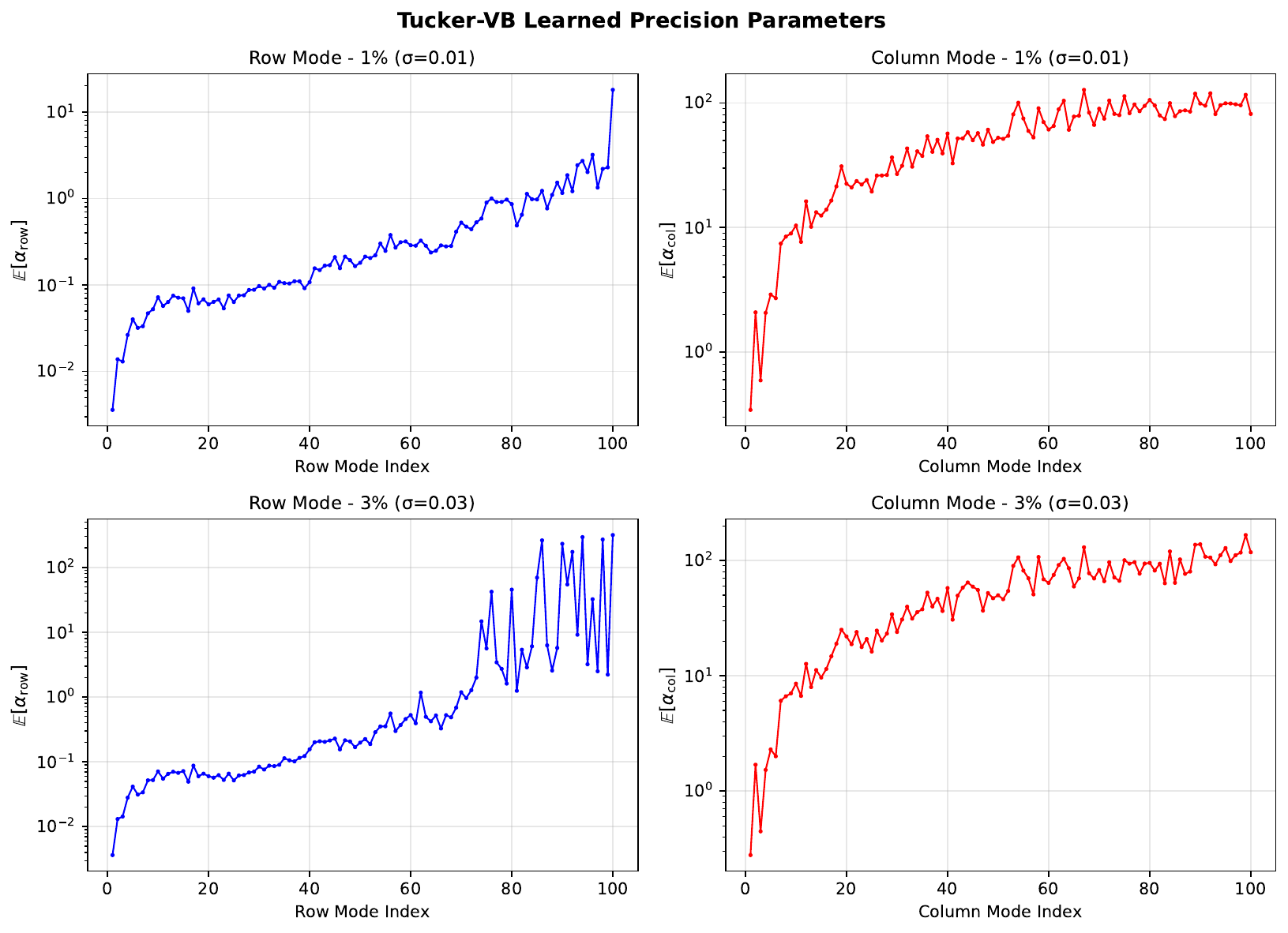}
	\caption{Precision parameters $\alpha_k$ learned by Tucker-VB. Left column: 
		row-direction precision $\alpha_k^{\mathrm{row}}$; right column: column-direction 
		precision $\alpha_l^{\mathrm{col}}$. Top row: $\sigma = 0.01$; bottom row: 
		$\sigma = 0.03$.}
	\label{fig:ard_params}
\end{figure*}

Figure~\ref{fig:ard_params} shows the precision parameters $\{\alpha_k^{\mathrm{row}}\}_{k=1}^{100}$ and $\{\alpha_l^{\mathrm{col}}\}_{l=1}^{100}$ learned by Tucker-VB.

\textbf{(1) Row versus column precision.}
Row-direction parameters $\alpha_k^{\mathrm{row}}$ span roughly $10^0$ to $10^3$; column-direction parameters $\alpha_l^{\mathrm{col}}$ span $10^{-1}$ to $10^2$---a gap of 1--2 orders of magnitude, with row regularization stronger. This matches the blur setup: row-direction blur $\sigma_{\mathrm{row}} = 1.3$ exceeds column-direction blur $\sigma_{\mathrm{col}} = 0.7$, so singular values decay faster and ill-conditioning is worse along rows, demanding stronger regularization. Tucker-VB learns this asymmetry from the data.

\textbf{(2) Precision versus mode index.}
Precision parameters rise with mode index in both directions. Within the Tucker framework, lower indices correspond to larger singular values (low-frequency content), higher indices to smaller singular values (high-frequency content). High-frequency components, tied to small singular values, amplify noise more during inversion and need heavier regularization. The increasing trend in Figure~\ref{fig:ard_params} shows Tucker-VB assigning weaker penalties to low frequencies and stronger penalties to high frequencies automatically.

Comparing top and bottom rows, overall regularization levels rise as noise increases from $\sigma = 0.01$ to $\sigma = 0.03$. For row direction, the peak $\alpha_k^{\mathrm{row}}$ stays near $10^{2.5}$ at both noise levels, but intermediate modes ($k = 40$--$80$) see noticeably higher values at $\sigma = 0.03$. Tucker-VB senses the noise increase and strengthens regularization accordingly.
Figure~\ref{fig:ard_params} clarifies why Tucker-VB outperforms single-parameter methods. Traditional methods apply one $\alpha$ uniformly across all modes, blind to row/column asymmetry and frequency-dependent requirements. Tucker-VB learns separate $\alpha_k^{\mathrm{row}}$ and $\alpha_l^{\mathrm{col}}$ for each coefficient, tailoring regularization strength at a fine granularity. This adaptive strategy underlies Tucker-VB's edge on anisotropic degradation problems.

\subsection{3D Heat Conduction Inverse Problem}

The backward heat conduction problem~\cite{XJAZ91333BE1CC346A9D9C5D2B51C85CD710,SJESE52000453C1771E1B5EB88DF44BD663E} serves as a benchmark PDE inverse problem. Consider an anisotropic heat equation~\cite{SJOXC0CBCB8C5636DC5E526D03DB38A6EAF6} on $\Omega = [0,1]^3$:
\begin{equation*}
	\frac{\partial u}{\partial t} = \kappa_x \frac{\partial^2 u}{\partial x^2} 
	+ \kappa_y \frac{\partial^2 u}{\partial y^2} + \kappa_z \frac{\partial^2 u}{\partial z^2}, 
	\quad (x,y,z) \in \Omega, \; t > 0
\end{equation*}
with temperature $u(x,y,z,t)$ and directional diffusivities $\kappa_x, \kappa_y, \kappa_z > 0$. Unequal diffusivities ($\kappa_x \neq \kappa_y \neq \kappa_z$) produce direction-dependent heat propagation. Periodic boundary conditions apply; the initial condition is $u(x,y,z,0) = f(x,y,z)$.
Fourier expansion gives:
\begin{equation*}
	u(x,y,z,t) = \sum_{l,m,n=-\infty}^{\infty} \hat{f}_{lmn} \cdot e^{-\mu_{lmn} t} 
	\cdot \phi_{lmn}(x,y,z)
\end{equation*}
with frequency indices $l, m, n \in \mathbb{Z}$, basis functions $\phi_{lmn}(x,y,z) = e^{2\pi i(lx + my + nz)}$. Fourier coefficients $\hat{f}_{lmn} = \langle f, \phi_{lmn} \rangle$, and eigenvalues $\mu_{lmn} = \kappa_x (2\pi l)^2 + \kappa_y (2\pi m)^2 + \kappa_z (2\pi n)^2$.
The goal is to recover $f(x,y,z)$ from noisy terminal observations $g(x,y,z) = u(x,y,z,T) + \epsilon(x,y,z)$. Directions with larger $\kappa$ suffer faster eigenvalue decay and worse conditioning, calling for stronger regularization.

Setup: $48 \times 48 \times 48$ grid ($N = 110{,}592$ unknowns); $\kappa_x = 0.01$, $\kappa_y = 0.005$, $\kappa_z = 0.02$; $T = 0.1$. The initial field combines three Gaussian sources:
\begin{equation*}
	f(x,y,z) = \sum_{k=1}^{3} A_k \exp\left(-\frac{(x - c_{x,k})^2 + (y - c_{y,k})^2 
		+ (z - c_{z,k})^2}{2\sigma_k^2}\right)
\end{equation*}
with intensities $A_k$, centers $(c_{x,k}, c_{y,k}, c_{z,k})$, and widths $\sigma_k$. Noise $\epsilon \sim \mathcal{N}(0, \sigma^2 I)$ is added at $\sigma = 0.01, 0.05, 0.1$ (1\%, 5\%, 10\%).

Table~\ref{tab:full_comparison} reports PSNR for five methods. Tucker-VB leads at all noise levels.
\begin{table*}[!ht]
	\centering
	\caption{Comparison of regularization methods for 3D backward heat equation inverse problem}
	\begin{tabular}{l c c c c c c}
		\toprule
		& \multicolumn{3}{c}{PSNR (dB)} & \multicolumn{3}{c}{Tucker-VB Gain (dB)} \\
		\cmidrule(lr){2-4} \cmidrule(lr){5-7}
		Method & 1\% noise & 5\% noise & 10\% noise & 1\% noise & 5\% noise & 10\% noise \\ \midrule
		L-curve   & 36.74 & 33.38 & 30.99 & +12.81 & +8.01  & +7.88  \\
		GCV       & 36.97 & 30.38 & 27.26 & +12.58 & +11.01 & +11.61 \\
		UPRE      & 36.95 & 30.39 & 27.31 & +12.60 & +11.00 & +11.56 \\
		DP        & 42.72 & 35.41 & 31.42 & +6.83  & +5.98  & +7.45  \\
		Tucker-VB & \textbf{49.55} & \textbf{41.39} & \textbf{38.87} & --- & --- & --- \\
		\bottomrule
	\end{tabular}
	\label{tab:full_comparison}
\end{table*}
Against DP, Tucker-VB gains 6.83, 5.98, and 7.45~dB at noise levels 1\%, 5\%, 10\%---about 6.75~dB on average. L-curve lags by 12.81~dB at low noise, 8~dB at higher levels. GCV and UPRE trail by over 11~dB once noise reaches 5\%. On average, Tucker-VB improves on DP by 6.75~dB and on L-curve by 9.57~dB. Per-component regularization and built-in noise estimation drive these gains.

\begin{figure*}
	\centering
	\includegraphics[width=\textwidth]{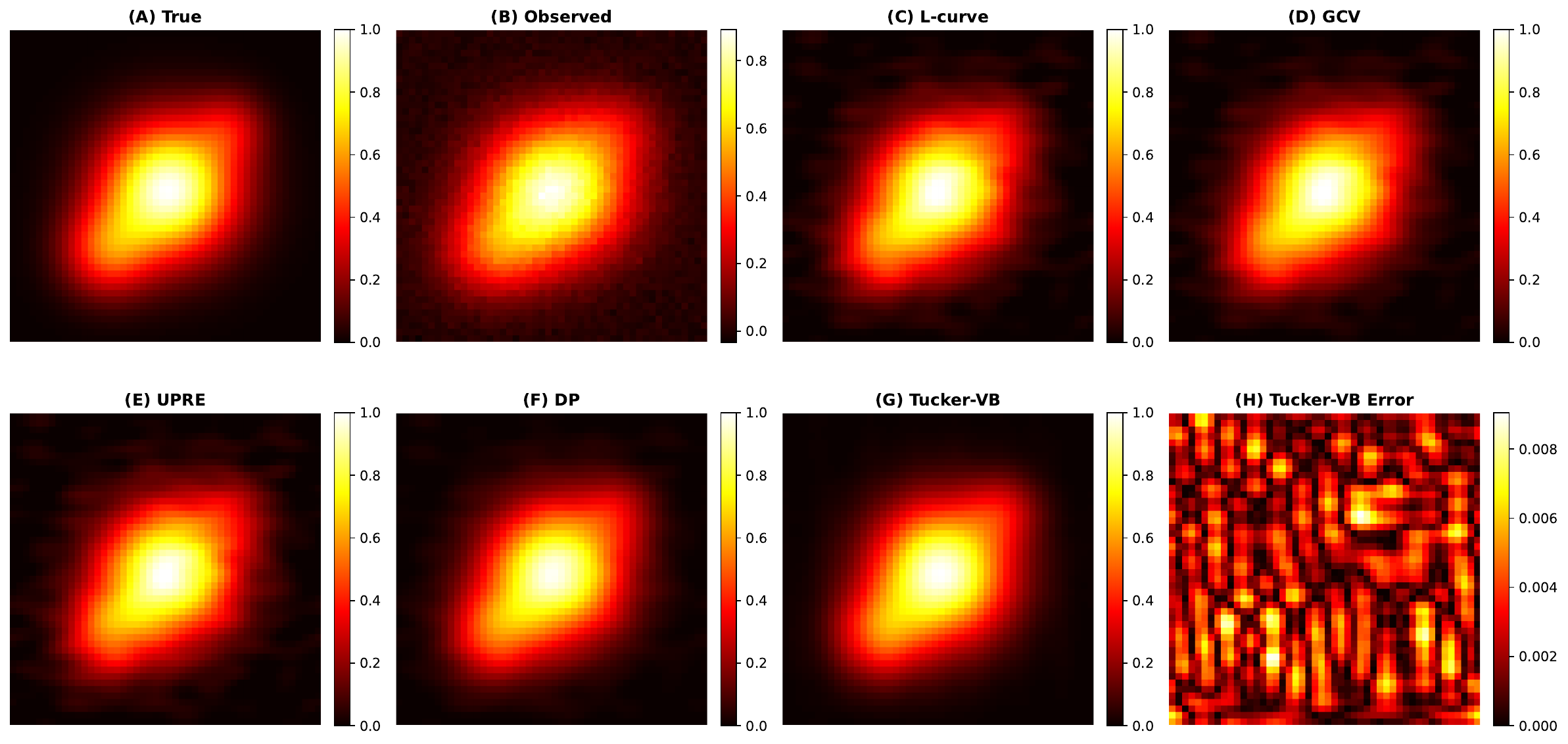}
	\caption{Reconstruction results for 3D backward heat equation at $\sigma = 0.01$ 
		(1\% noise). (A) True initial temperature field; (B) Noisy observation; 
		(C) L-curve reconstruction; (D) GCV reconstruction; (E) UPRE reconstruction; 
		(F) DP reconstruction; (G) Tucker-VB reconstruction; (H) Tucker-VB absolute error map.}
	\label{fig:reconstruction_1pct}
\end{figure*}

\begin{figure*}
	\centering
	\includegraphics[width=\textwidth]{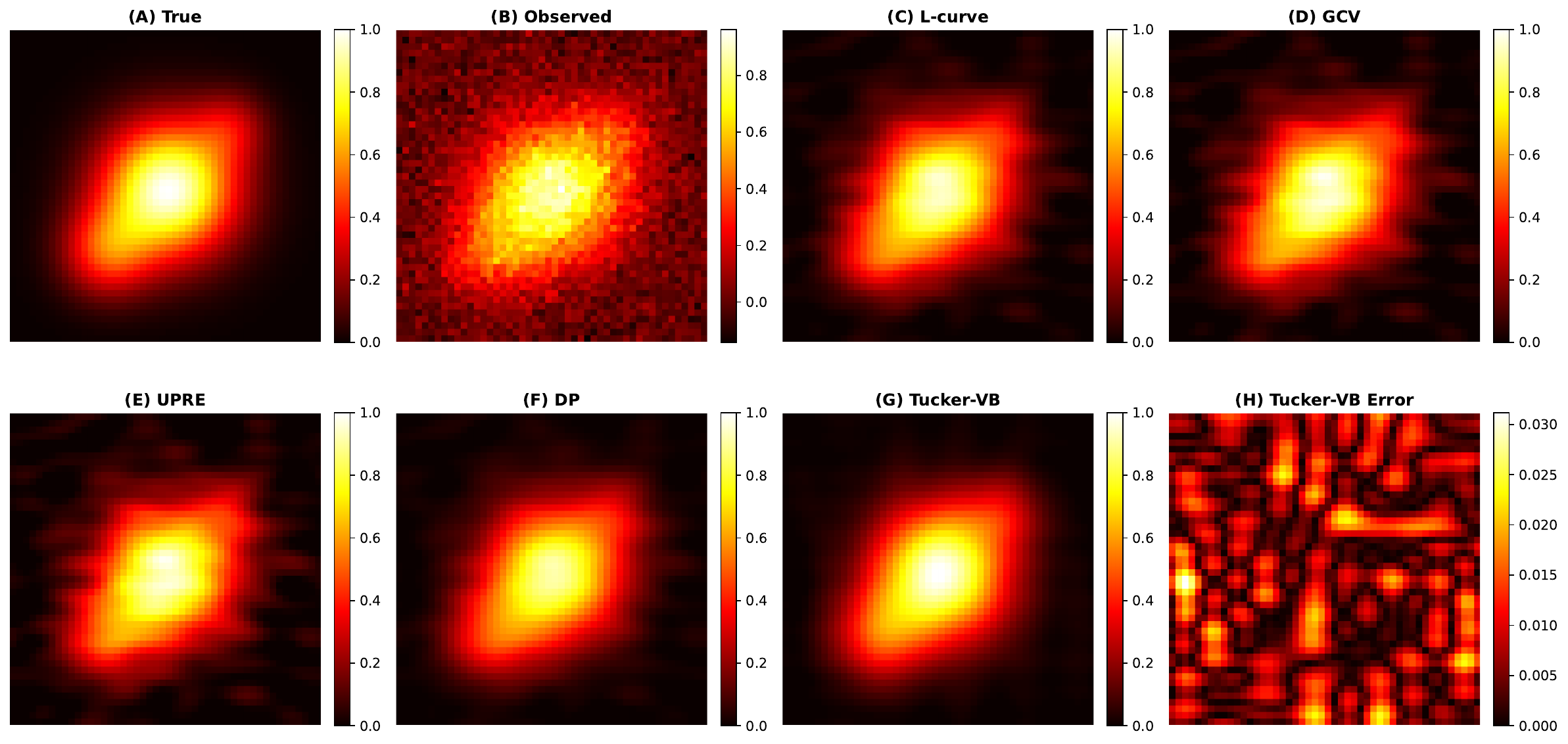}
	\caption{Reconstruction results for 3D backward heat equation at $\sigma = 0.05$ 
		(5\% noise). (A) True initial temperature field; (B) Noisy observation; 
		(C) L-curve reconstruction; (D) GCV reconstruction; (E) UPRE reconstruction; 
		(F) DP reconstruction; (G) Tucker-VB reconstruction; (H) Tucker-VB absolute error map.}
	\label{fig:reconstruction_5pct}
\end{figure*}

\begin{figure*}
	\centering
	\includegraphics[width=\textwidth]{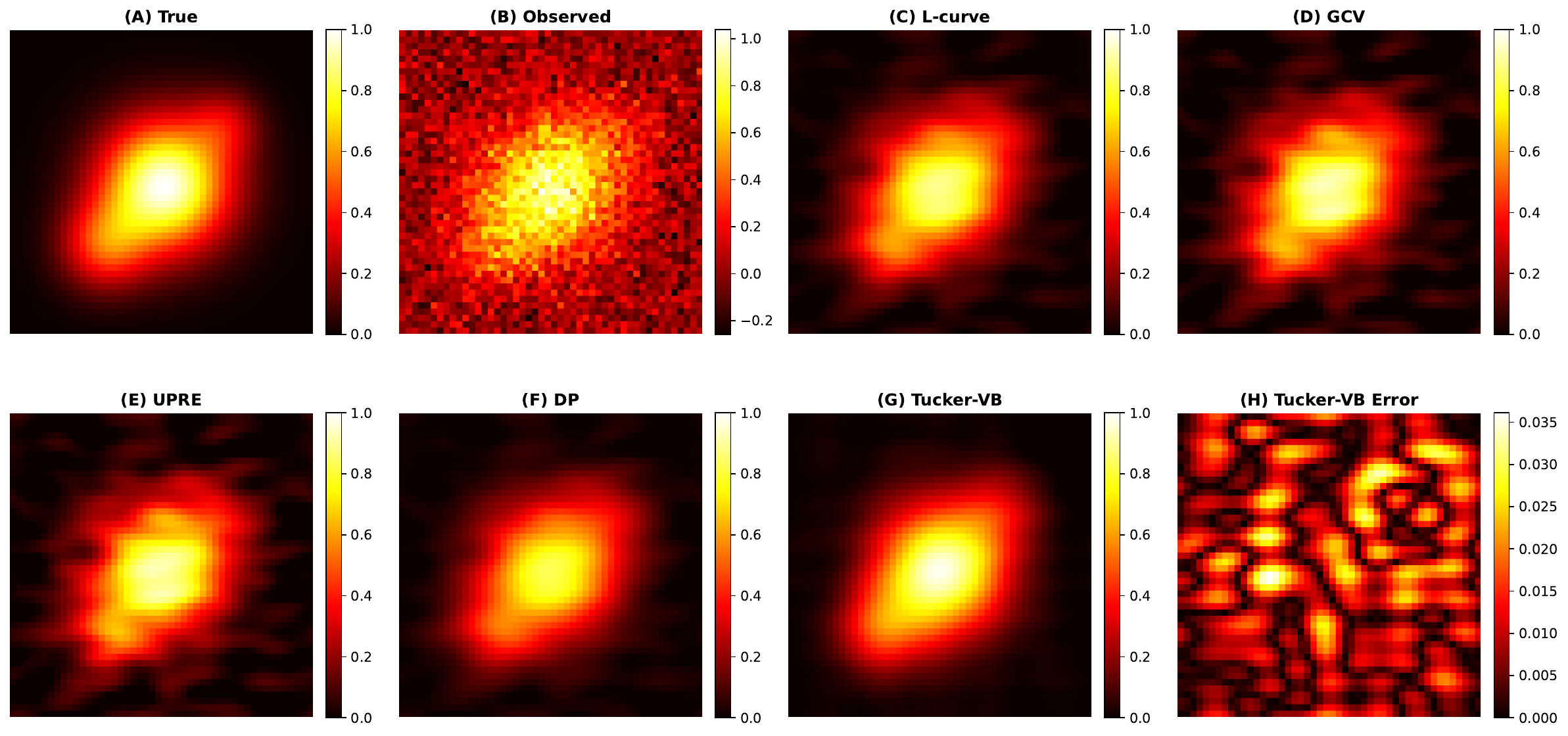}
	\caption{Reconstruction results for 3D backward heat equation at $\sigma = 0.1$ 
		(10\% noise). (A) True initial temperature field; (B) Noisy observation; 
		(C) L-curve reconstruction; (D) GCV reconstruction; (E) UPRE reconstruction; 
		(F) DP reconstruction; (G) Tucker-VB reconstruction; (H) Tucker-VB absolute error map.}
	\label{fig:reconstruction_10pct}
\end{figure*}

Figures~\ref{fig:reconstruction_1pct}--\ref{fig:reconstruction_10pct} show reconstructions. We plot the central $z$-slice ($z_0 = 24$) since 3D fields are awkward to display. All temperature maps use the same ``hot'' colormap on $[0,1]$; error maps (panel H) rescale automatically.

\begin{figure*}
	\centering
	\includegraphics[width=\textwidth]{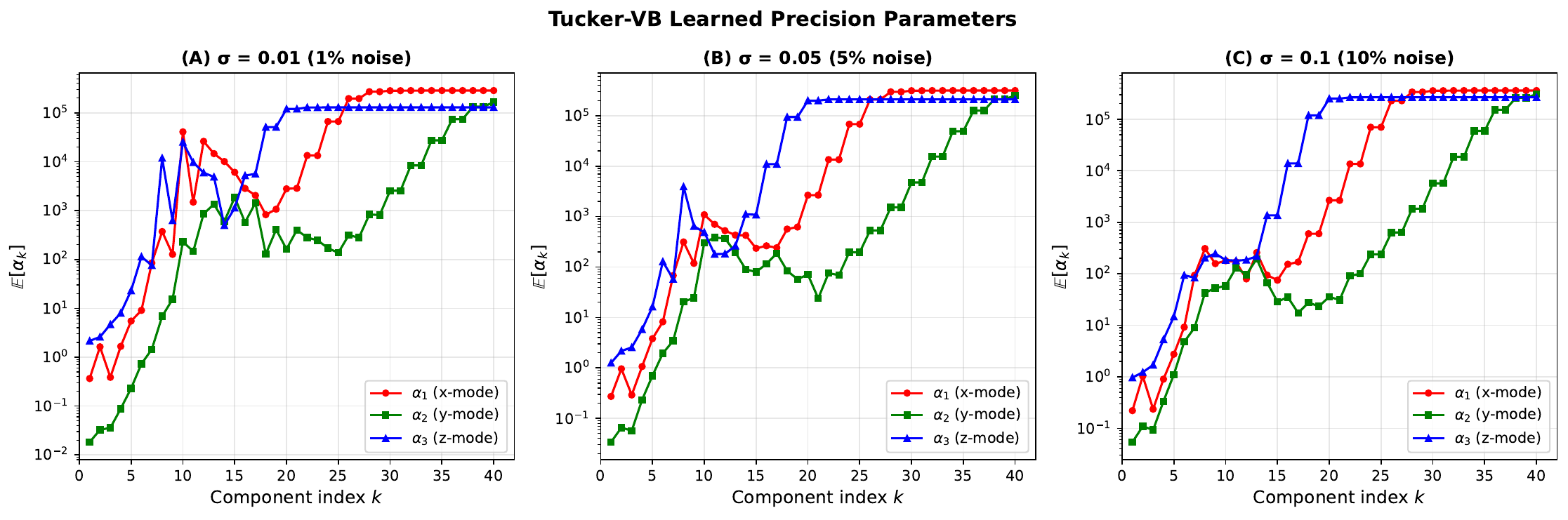}
	\caption{Precision parameters learned by Tucker-VB at different noise levels. 
		(A) $\sigma = 0.01$; (B) $\sigma = 0.05$; (C) $\sigma = 0.1$.}
	\label{fig:ard_precision}
\end{figure*}

Tucker-VB (panel G) recovers source positions, shapes, intensities, and boundaries most faithfully at all noise levels. At 10\% noise (Figure~\ref{fig:reconstruction_10pct}(G)), the main temperature features remain intact. GCV and UPRE develop artifacts as noise grows. Panel (H) maps absolute error $|f_{\mathrm{true}} - \hat{f}|$.
Figure~\ref{fig:ard_precision} plots precision parameters $\alpha$ learned at each noise level.
The curves show precision versus component index along each axis. Low $\alpha$ marks signal-rich components; high $\alpha$ marks noise-dominated ones. Growth rates follow $\kappa_z > \kappa_x > \kappa_y$: the $z$-curve climbs fastest, then $x$, then $y$. Tucker-VB has learned the physical anisotropy from data alone.

\section{Conclusion}

This paper develops a variational Bayesian method under Tucker low-rank constraints for regularization parameter selection in high-dimensional inverse problems. The approach overcomes the dimensionality bottleneck that limits standard variational Bayes in large-scale settings.
The main idea is to combine tensor low-rank structure with Bayesian inference. Tucker decomposition not only compresses high-dimensional data but also shifts variational inference from the original space to a low-dimensional core tensor space, bypassing the $O(n^3)$ complexity and $O(n^2)$ storage costs of conventional variational Bayes. We derive a complete mathematical framework for Bayesian models in Tucker subspaces. The multi-hyperparameter prior extends global regularization to mode-adaptive regularization, suited for anisotropic problems.
Four sets of numerical experiments test the method. Tucker-VB runs orders of magnitude faster than direct VB and scales to problem sizes where direct VB fails. Reconstruction accuracy matches or exceeds all baselines across test problems and noise levels: 20\% improvement on integral equations, 0.73--2.09~dB on image deblurring, 6.75~dB on heat conduction. The learned precision parameters track the physical structure of each problem, confirming that mode-wise regularization works as intended.
Tucker-VB bridges the gap between variational Bayes theory and large-scale computation. With multi-dimensional data now common in medical imaging, remote sensing, and scientific computing, the method should find use in a range of applications.


\end{document}